\documentclass[letterpaper,journal]{IEEEtran}

\usepackage{amsmath,amsfonts}
\usepackage{algorithm}
\usepackage{algorithmic}
\usepackage{graphicx}
\usepackage{array}
\usepackage{pifont}
\usepackage[normalem]{ulem}
\usepackage[caption=false,font=normalsize,labelfont=sf,textfont=sf]{subfig}
\usepackage{textcomp}
\usepackage{stfloats}
\usepackage{url}
\usepackage{verbatim}
\usepackage{cite}
\usepackage{booktabs}
\usepackage{tabularx}
\usepackage{makecell}
\newcommand{\cmark}{\ding{51}}
\newcommand{\xmark}{\ding{55}}

\newcolumntype{Y}{>{\centering\arraybackslash}X}

\begin{document}
%
\title{KG-ASG: Collision-Knowledge-Guided Closed-Loop Adversarial Scenario Generation With Primary-Support Attribution}
%
%
%
\author{Cheng Wang, Chen Xiong, Ziwen Wang, Yuchen Zhou, \textit{Graduate Student Member IEEE}, and Qiang Liu%
\thanks{This work was supported by the Science and Technology Innovation Key R\&D Program of Chongqing (CSTB2023TIAD-STX0030), the National Key R\&D Program of China (Grant No. 2024YFB4303400), Natural Science Foundation of Guangdong Province, China (Grant No. 2025A1515010166), and the Shenzhen Fundamental Research Program, China (Grant No. JCYJ20240813151301003).}%
\thanks{Cheng Wang, Chen Xiong, Ziwen Wang, Yuchen Zhou and Qiang Liu are with the Guangdong Provincial Key Laboratory of Intelligent Transportation System, School of Intelligent Systems Engineering, Shenzhen Campus of Sun Yat-sen 
University, Shenzhen 518107, China (e-mail: wangch526@mail2.sysu.edu.cn; xiongch8@mail.sysu.edu.cn; wangzw27@mail2.sysu.edu.cn; zhouych37@mail2.sysu.edu.cn; liuq32@mail.sysu.edu.cn).}%

\thanks{Corresponding author: Qiang Liu, Chen Xiong.}}


%
%

\markboth{arXiv Preprint}%
{Shell \MakeLowercase{\textit{et al.}}: Bare Demo of IEEEtran.cls for IEEE Journals}
%



\maketitle

\begin{abstract}
Safety validation of autonomous driving systems requires not only high-risk scenario coverage, but also clear collision semantics, executable trajectories, and attributable multi-vehicle interactions. Existing safety-critical scenario generation methods often rely on low-level trajectory perturbations, collision-proxy optimization, or single-adversary search, which may produce adversarial samples with ambiguous collision causes or uncontrolled multi-vehicle collisions. This paper proposes KG-ASG, a collision-knowledge-guided closed-loop adversarial scenario generation framework with primary-support attribution. KG-ASG first constructs a structured collision knowledge base and trains a lightweight Collision Expert to infer the target collision mode, the unique primary adversary, support vehicles, and their interaction roles. Guided by this semantic prior, multi-vehicle adversarial generation is formulated as a primary-support process, where the primary adversary induces the main conflict and support vehicles shape the surrounding risk structure without becoming additional colliders. Rule, physical, interaction-safety, and single-collider constraints are imposed as hard gates to filter non-executable samples. To handle reactive ego behaviors, planner-controller feedback is further used for failure diagnosis, candidate re-ranking, and terminal refinement. Experiments on WOMD scenarios reconstructed in MetaDrive show that KG-ASG achieves a 92.60\% attack success rate under the Replay policy and maintains 92.40\% under joint rule and physical constraints. KG-ASG also improves Valid Primary Attack to 92.20\%, reduces multi-collision to 0.00\%, and obtains closed-loop recovery gains of 8.00, 7.20, and 14.00 percentage points under IDM, Cruise, and Expert controllers, respectively. These results demonstrate that collision-knowledge guidance and primary-support single-collider reasoning improve adversarial effectiveness, interpretability, and executability, providing a structured scenario-generation tool for autonomous driving safety validation in intelligent transportation systems.

\end{abstract}

\begin{IEEEkeywords}
Autonomous driving safety testing, adversarial scenario generation, collision knowledge, primary-support reasoning, closed-loop feedback.
\end{IEEEkeywords}

%
\IEEEpeerreviewmaketitle

\newpage

\section{Introduction}
%
%
%
%
\IEEEPARstart{S}{dafety} validation remains a central challenge at the intersection of intelligent transportation and artificial intelligence. Although autonomous driving systems have made substantial progress in perception, prediction, planning, and control, demonstrating their safety in open, continuous, and highly uncertain real-world traffic environments remains far from solved. Prior studies have shown that validating autonomous driving systems solely through accumulated naturalistic driving mileage suffers from prohibitive sample complexity and testing cost~\cite{liu2024curse}. Meanwhile, real-world crashes, disengagements, and failure cases continue to reveal non-negligible vulnerabilities of autonomous driving systems in complex interactions, atypical maneuvers, and long-tail conflict scenarios~\cite{abdelaty2024matched,yang2024guarantee,zhou2024realworldcrash}. Therefore, automatically constructing high-risk, interpretable, controllable, and executable adversarial testing scenarios has become an important research direction for autonomous driving safety validation~\cite{zhou2024evaluating,wei2025interactive,liu2024safetycriticalediting}. \par

\begin{figure}[t]
    \centering
    \includegraphics[width=0.85\columnwidth]{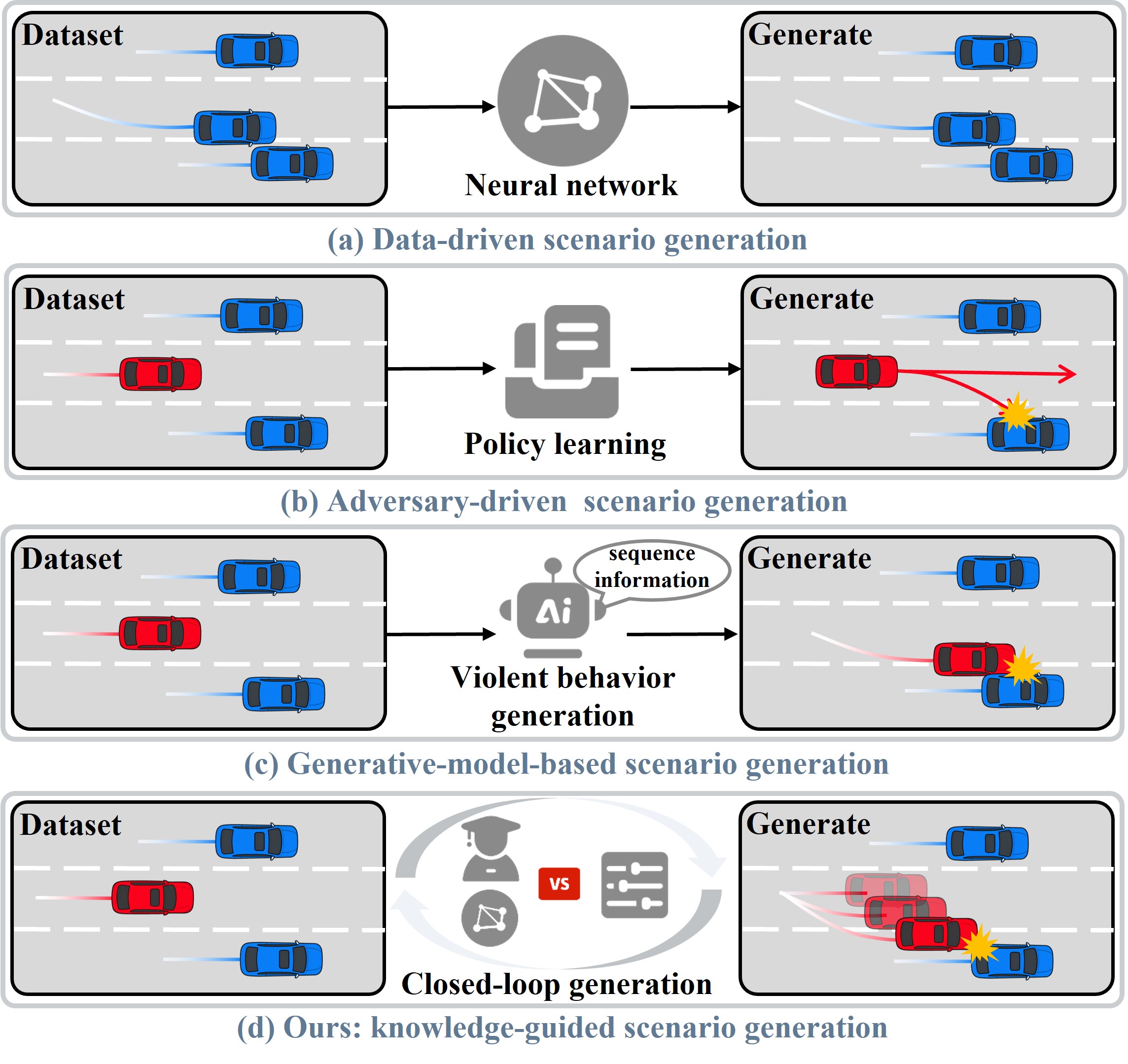}
    \caption{Comparison of scenario generation paradigms for autonomous driving safety testing. Existing methods can be broadly categorized into data-driven, adversary-driven, and generative-model-based scenario generation. In contrast, KG-ASG introduces a knowledge-guided closed-loop paradigm, where collision semantics and structured risk roles guide adversarial generation rather than relying solely on data imitation, policy perturbation, or generative priors.}
    \label{fig:scenario_paradigm}
\end{figure}

Existing safety-critical scenario generation methods can be broadly categorized into three paradigms, as illustrated in Fig.~\ref{fig:scenario_paradigm}. The first paradigm is data-driven safety-critical scenario learning and evaluation, which mines high-risk scenario distributions from real-world crashes, pre-crash trajectories, and naturalistic driving data for risk identification, safety evaluation, and test-sample extraction~\cite{zhou2024realworldcrash,zhou2024precrashptw,wei2025interactive}. These methods benefit from strong realism and statistical consistency, but they are primarily designed for scenario discovery and evaluation rather than proactive generation toward specific collision modes. The second paradigm is perturbation- and adversary-driven scenario generation, including accelerated evaluation, importance sampling, rule-based scenario construction, formal testing, reinforcement learning, multi-agent adversarial generation, and closed-loop simulation~\cite{yang2025adaptiveimportance,liu2024safetycriticalediting,zhang2025criticalrl,wang2026hspg}. These methods can actively amplify risky interactions and expose system-level weaknesses, but they often reduce ``risk'' to low-level collision proxies, scalar risk scores, or policy-failure events, making it difficult to explicitly represent collision semantics and complex multi-vehicle conflict structures. The third paradigm is generative-prior- and foundation-model-guided scenario generation, where traffic priors, diffusion models, world models, large language models, and multimodal foundation models are used to improve the naturalness, semantic controllability, and diversity of generated scenarios~\cite{xie2024advdiffuser,xu2025diffscene,guan2025worldmodel,zhang2024chatscene,mei2025llmattacker}. However, most existing methods still simplify scenario criticality as a collision probability, a trajectory-space risk proxy, or a reward preference. They rarely model explicitly which type of collision is targeted, which vehicle is responsible for triggering the main conflict, and how other vehicles shape the surrounding risk structure. As a result, generated scenarios may trigger collisions but remain difficult to interpret, reproduce, or attribute. \par

With the rapid development of large language models and multimodal foundation models, using them for traffic-scene understanding, traffic-knowledge reasoning, and scenario-generation guidance has become a promising direction~\cite{lan2025trajllm,chen2024asyncdriver,kuang2025trafficit}. Foundation models show strong capability in high-level context understanding, natural-language reasoning, and semantic abstraction. Nevertheless, directly using them to generate low-level continuous trajectories or planning-control actions remains challenging due to latency, computational cost, physical feasibility, and fine-grained control accuracy. We therefore argue that foundation models are better suited as high-level collision-knowledge and semantic-guidance modules, rather than as direct trajectory controllers. A more reliable design is to use a lightweight Collision Expert to understand collision modes, identify key interacting agents, and output structured generation guidance, while a specialized adversarial generation module and rule checker handle continuous trajectory selection, physical feasibility filtering, and closed-loop feedback refinement. \par

Motivated by this observation, this paper proposes KG-ASG , a knowledge-guided closed-loop adversarial scenario generation framework. KG-ASG follows the principle of first understanding collision semantics, then guiding adversarial generation, and finally refining scenarios through closed-loop feedback. Specifically, KG-ASG first builds a structured collision knowledge base from crash typology, real-world crash and disengagement cases, autonomous-driving scenario engineering knowledge, and safety-critical scenario generation experience. To avoid relying on short-horizon geometric heuristics alone, KG-ASG uses UIRAM-based risk assessment to pre-screen candidate interacting vehicles, thereby incorporating interaction relevance, intention uncertainty, collision likelihood, and potential consequence into the candidate selection stage~\cite{wang2026uiram}. Based on the resulting candidate set, a lightweight Collision Expert predicts the target collision mode, the unique primary adversary, support vehicles, role assignments, and critical spatiotemporal guidance. KG-ASG then transforms adversarial scenario generation from low-level proxy-driven perturbation into knowledge-guided primary-support generation: the primary adversary is responsible for inducing the main conflict or near-miss, while support vehicles compress the ego vehicle's feasible avoidance space, adjust interaction timing, or shape the local risk structure without becoming additional colliders. Road topology, traffic signal, dynamic feasibility, interaction safety, and single-collider constraints are further imposed as hard filters. Finally, generated scenarios are executed with reactive planner-controller policies, and the resulting ego trajectories, near-miss behaviors, braking responses, and failure types are used to retry, re-rank, and terminally refine adversarial candidates. \par

The main contributions of this work are summarized as follows:
\begin{itemize}
    \item We propose a structured collision-knowledge modeling method for adversarial scenario generation, which unifies collision modes, risk semantics, and generation guidance to provide interpretable semantic priors for long-tail risk reasoning and targeted scenario generation.

    \item We develop a Collision-Expert-guided primary-support adversarial generation framework. By coupling high-level collision semantics with low-level candidate trajectory generation, the proposed method shifts adversarial generation from single proxy-driven collision search to a structured process in which the primary adversary induces the main conflict and support vehicles shape the surrounding risk structure.

    \item We introduce a closed-loop regeneration mechanism that integrates rule constraints and planner-controller feedback. Under road, signal, dynamic, interaction-safety, and single-collider constraints, KG-ASG uses feedback-driven retry, candidate re-ranking, and terminal refinement to recover attacks avoided by reactive controllers while filtering unrealistic or non-executable adversarial samples.
\end{itemize}

Section 2 reviews related work and summarizes the research gaps. Section 3 formulates the knowledge-guided adversarial scenario generation problem. Section 4 presents the proposed KG-ASG framework, including collision knowledge modeling, primary-support trajectory generation, and rule-constrained closed-loop regeneration. Section 5 reports experimental results and ablation studies. Section 6 concludes this paper and discusses future work. \par

\section{Related Work}

This section reviews safety-critical and adversarial scenario generation for autonomous driving from three perspectives: data-driven safety-critical scenario learning and evaluation, perturbation- and adversary-driven hazardous scenario generation, and generative-prior- or foundation-model-guided scenario generation. We focus on how existing studies define scenario criticality, whether they provide explicit collision-semantic attribution, and how they handle multi-vehicle interaction, rule executability, and closed-loop planner-controller feedback. \par

\subsection{Data-Driven Safety-Critical Scenario Learning and Evaluation}

Data-driven methods provide an empirical foundation for autonomous-driving safety validation by learning risk distributions from real-world crashes, pre-crash trajectories, disengagement events, and naturalistic driving data. Structured pre-crash typologies remain important references for organizing accident mechanisms and safety-critical interaction patterns~\cite{najm2007precrash}. Recent studies further show that safety validation cannot rely solely on naturalistic mileage accumulation due to the rarity of safety-critical events~\cite{liu2024curse}. Real-world crash and pre-crash analyses have compared autonomous and human-driven vehicle crashes, investigated autonomous-vehicle behavior in crash scenarios, and evaluated safety performance through pre-crash trajectories of powered two-wheelers~\cite{abdelaty2024matched,zhou2024realworldcrash,zhou2024precrashptw}. These studies reveal that autonomous driving systems remain vulnerable in complex interactions, atypical maneuvers, and long-tail conflict situations. \par

Recent work has moved from direct crash replay toward pre-crash generalization and scenario-family construction. Cut-in pre-crash generalization, master-scenario mining, improved adaptive stress testing, roundabout scenario generation, cascaded safety analysis, and interaction-coding-based construction enlarge the coverage of typical high-risk interactions~\cite{li2024cutin,li2024masterscenarios,zhou2025improvedast,ren2025roundabouts,sun2025cascaded,chang2025interactioncoding}. Reinforcement-learning-based interactive critical scenario generation further converts in-depth crash knowledge into more challenging test samples~\cite{wei2025interactive}. Nevertheless, data-driven methods are still constrained by crash-data scale, annotation quality, and long-tail coverage. They provide valuable risk priors and accident-mechanism clues, but they rarely specify which vehicle should be responsible for the main conflict and which vehicles should only shape the surrounding risk structure. This motivates using structured collision knowledge as a semantic prior for targeted and attributable adversarial generation. \par

\subsection{Perturbation- and Adversary-Driven Hazardous Scenario Generation}

Perturbation- and adversary-driven methods actively modify traffic participants, trajectories, or scene parameters to amplify risky interactions and expose weaknesses of autonomous driving systems. Early studies use accelerated evaluation, importance sampling, formal testing, and rule-based scenario construction to improve rare-event sampling efficiency and test verifiability~\cite{yang2025adaptiveimportance,zhou2026fullscenario}. Reinforcement-learning and multi-agent adversarial methods further formulate safety-critical scenario generation as an interaction process, where adversarial agents induce collisions, near misses, emergency braking, or planning failures~\cite{liu2024safetycriticalediting,zhang2025criticalrl,wei2025interactive}. Recent proactive or behavior-guided studies, including HSPG, emergency lane-change simulation, and MJTG, show that proactive actor selection, typical hazardous maneuvers, and multi-vehicle joint trajectory generation can improve the coverage of complex long-tail risks~\cite{wang2026hspg,xiong2026emergency,tian2025mjtg}. \par

Candidate-trajectory-based and adversarial re-ranking methods provide strong baselines for interaction-level scenario generation. CAT searches adversarial opponent trajectories in a candidate trajectory space and evaluates their effects through ego rollouts~\cite{zhang2023cat}. KING, AdvTrajOpt, SEAL, GOOSE, and SAGE further explore different mechanisms for improving attack success, trajectory plausibility, map compliance, and distributional realism~\cite{hanselmann2022king,zhang2022advtrajopt,stoler2025seal,ransiek2024goose,nie2026sage}. However, most existing adversary-driven methods still define criticality through collision occurrence, scalar rewards, or low-level risk proxies. In multi-vehicle scenarios, they often do not explicitly distinguish the vehicle responsible for the final conflict from vehicles that only provide blocking, pressure, or timing-shaping effects. This may inflate attack success through ambiguous attribution or uncontrolled multi-collision, reducing interpretability and reproducibility. \par

\subsection{Generative-Prior- and Foundation-Model-Guided Scenario Generation}

Generative-prior-based methods aim to improve scenario realism, diversity, and controllability by learning traffic distributions, trajectory priors, and interaction patterns from data. Diffusion-based methods such as AdvDiffuser and DiffScene synthesize safety-critical scenarios through guided generative processes while preserving plausible trajectory distributions~\cite{xie2024advdiffuser,xu2025diffscene}. World-model-based scene generation further models pre-crash evolution and future scene dynamics, providing generative representations for accident anticipation and risk scenario synthesis~\cite{guan2025worldmodel}. These methods mitigate unnatural behaviors and distribution shifts commonly observed in direct adversarial perturbation, but they usually emphasize distributional plausibility or future-scene generation rather than explicit modeling of collision modes, primary colliders, support roles, and single-collider attribution. \par

Large language models and multimodal foundation models have recently been introduced into traffic-scene understanding, scenario generation, perception testing, and planning guidance. ChatScene uses knowledge-enabled reasoning for safety-critical scenario generation, while LLM-Attacker uses multi-agent large language models to identify attack targets and enhance closed-loop adversarial scenario generation through system feedback~\cite{zhang2024chatscene,mei2025llmattacker}. Related studies explore foundation models for trajectory prediction, traffic-scene understanding, asynchronous planning, context-aware motion prediction, concrete simulation scenario generation, dynamic prompting, and safety-critical curriculum learning~\cite{lan2025trajllm,kuang2025trafficit,chen2024asyncdriver,zheng2024contextmotion,li2025automatingconcrete,danso2025automated,yang2025trajectoryllm,zhang2026dynamicprompting,sheng2026curricuvlm}. These works demonstrate strong high-level semantic reasoning, but directly using foundation models to generate continuous trajectories or low-level control actions remains limited by latency, physical feasibility, and fine-grained control accuracy. Moreover, primary-support role separation, support no-collision constraints, and single-collider attribution are rarely enforced. \par

\subsection{Summary}

In summary, existing studies have advanced autonomous-driving safety-critical scenario generation through crash-data analysis, pre-crash typology, scenario-family construction, trajectory perturbation, reinforcement learning, candidate-trajectory search, adversarial re-ranking, multi-vehicle joint generation, generative modeling, and foundation-model reasoning. However, three limitations remain insufficiently addressed. First, many methods still define scenario criticality mainly through collision occurrence, scalar risk scores, adversarial rewards, or trajectory-distribution shifts, with limited explicit modeling of collision modes and accident-mechanism priors. Second, multi-vehicle generation often lacks primary-support role separation and single-collider attribution constraints, making it difficult to distinguish the main conflict vehicle from vehicles that only shape the surrounding risk structure. Third, rule executability and planner-controller feedback are often handled as separate post-processing or evaluation steps, rather than being integrated into a unified adversarial generation loop. \par

KG-ASG addresses these gaps by combining structured collision knowledge, Collision-Expert-guided primary-support assignment, role-aware candidate trajectory re-ranking, hard rule and physical constraints, and planner-controller-feedback-driven regeneration. The proposed framework does not merely aim to increase collision frequency. Instead, it seeks to generate high-risk scenarios that are executable, attributable, and semantically interpretable. \par

\section{Problem Formulation}

This paper focuses on adversarial scenario generation for safety testing of autonomous-driving planner-controller systems. Given an initial scenario reconstructed from real-world traffic logs, the objective is to generate high-risk traffic scenarios that can expose potential vulnerabilities of the planner-controller while preserving traffic semantic plausibility, kinematic executability, and rule consistency. The generated scenarios are expected to induce challenging behaviors such as near misses, emergency braking, failed avoidance, or collisions. Unlike methods that rely only on low-level collision heuristics, scalar risk proxies, or single-opponent perturbations, we formulate this problem as a hierarchical generation task driven by scene representation, collision-knowledge reasoning, primary-support actor selection, conditional trajectory generation, and planner-controller closed-loop feedback. The key is not to blindly search for trajectories that collide with the ego vehicle, but to first identify the potential collision mode, key interacting vehicles, and their risk roles, and then convert them into structured guidance executable by the adversarial generation module.

\subsection{Scene Representation}

Consider a traffic scene composed of an ego vehicle \(o_0\) and \(N\) surrounding traffic participants \(\{o_1,\ldots,o_N\}\). Let \(T_{\mathrm{obs}}\) denote the observation horizon and \(T_{\mathrm{pred}}\) denote the generation horizon. For any vehicle \(o_i\), where \(i\in\{0,\ldots,N\}\), its historical state sequence is defined as
\begin{equation}
S_i=\left\{s_i^t\right\}_{t=-T_{\mathrm{obs}}+1}^{0}.
\end{equation}

The single-frame state vector is defined as
\begin{equation}
s_i^t=
\left[
x_i^t,\,
y_i^t,\,
v_i^t,\,
a_i^t,\,
\theta_i^t,\,
d_i,\,
c_i
\right].
\end{equation}

Here, \((x_i^t,y_i^t)\) denotes the position of vehicle \(o_i\) in the ego-centric coordinate system; \(v_i^t\), \(a_i^t\), and \(\theta_i^t\) denote its velocity, acceleration, and heading angle, respectively; \(d_i\) denotes the vehicle size; and \(c_i\) denotes the semantic category of the vehicle. The static environment is represented by a high-definition map \(M\), including lane topology, road boundaries, traffic signals, stop lines, drivable areas, and right-of-way relations. The complete traffic scene is therefore represented as
\begin{equation}
x=
\left(
M,\,
S_0,\,
S_{1:N}
\right).
\end{equation}

In this formulation, \(x\) denotes the current traffic scene state, \(S_0\) denotes the historical state sequence of the ego vehicle \(o_0\), and \(S_{1:N}=\{S_1,\ldots,S_N\}\) denotes the historical state sequences of surrounding vehicles. Instead of applying unconstrained generation to all surrounding vehicles, this paper first identifies the key participants that form major risk interactions with the ego vehicle and then performs targeted adversarial generation on them.

\subsection{Collision Knowledge and Interactive Actor Modeling}

Real-world traffic accidents are rarely caused by arbitrary trajectory perturbations. Instead, they usually exhibit explicit scene semantics, participant relations, and triggering behaviors. To introduce accident-mechanism priors, this paper constructs a collision knowledge base \(K\) and uses a Collision Expert to perform high-level semantic reasoning over the current scene. Let the Collision Expert be denoted by \(f_{\phi}\), where \(\phi\) represents the model parameters. Given the scene state \(x\) and the knowledge base \(K\), the expert outputs structured generation guidance:
\begin{equation}
g
=
f_{\phi}(x,K)
=
\left(
m^*,\,
o_{i^*},\,
B^*,\,
R_q,\,
u
\right).
\end{equation}

Here, \(g\) denotes the structured generation guidance; \(m^*\) denotes the target collision mode that is most relevant and adversarially valuable in the current scene; \(o_{i^*}\) denotes the unique primary adversary; \(B^*\) denotes the support vehicle set; \(R_q\) denotes the support-role set, such as blocker or rear-pressure vehicle; and \(u\) denotes the conditional information required for subsequent generation, including the target conflict region, critical time window, and behavior priority.

To avoid selecting only a single high-risk vehicle while ignoring local interaction structures, we formulate key actor selection as a small-scale primary-support combination problem. Let the candidate vehicle set be
\begin{equation}
O(x)=\{o_1,\ldots,o_N\}.
\end{equation}

The final set of vehicles participating in adversarial generation is defined as
\begin{equation}
V^*
=
\{o_{i^*}\}\cup B^*,
\qquad
|B^*|\leq 2.
\end{equation}

Here, \(V^*\) denotes the final adversarial vehicle set. The vehicle \(o_{i^*}\) is the only vehicle allowed to induce the main collision or the main near-miss. The support vehicles in \(B^*\) do not include the primary adversary; they are only allowed to shape the local risk structure and are not allowed to collide directly with the ego vehicle \(o_0\). This setting transforms ``multi-vehicle joint attack'' into a structured formulation in which the primary adversary is responsible for the main conflict and support vehicles are responsible for risk shaping, thereby improving the interpretability of collision attribution.

\subsection{Conditional Adversarial Scenario Generation}

After obtaining the high-level guidance \(g\), we formulate adversarial scenario generation as a conditional trajectory generation problem. Let \(T(o_j)\) denote the candidate trajectory space of vehicle \(o_j\). The adversarial scenario generated from scene \(x\) is represented as
\begin{equation}
A(x)
=
\left\{
(o_j,\tau_j)
\mid
o_j\in V^*,\,
\tau_j\in T(o_j)
\right\}.
\end{equation}

Here, \(A(x)\) denotes the generated adversarial scenario and \(\tau_j\) denotes the future trajectory of vehicle \(o_j\). Unlike unguided search, the generation process is jointly conditioned on the target collision mode, primary-support roles, critical conflict region, and time window. Accordingly, adversarial scenario generation is formulated as the following constrained optimization problem:
\begin{equation}
A^*
=
\arg\max_{A\in\Omega(x),\,C(A)=1}
J(A;x,\pi_c).
\end{equation}

Here, \(A^*\) denotes the selected adversarial scenario; \(\Omega(x)\) denotes the candidate adversarial scenario space under scene \(x\); \(J(A;x,\pi_c)\) denotes the adversarial objective with respect to the planner-controller \(\pi_c\); and \(C(A)=1\) indicates that the scenario satisfies hard constraints, including road topology, traffic signals, dynamic feasibility, and interaction safety.

To ensure clear collision attribution, we further impose the single-collider constraint:
\begin{equation}
\sum_{o_j\in V^*}
I
\left\{
\mathrm{Collide}(o_j,o_0)
\right\}
\leq 1,
\end{equation}
\begin{equation}
I
\left\{
\mathrm{Collide}(o_j,o_0)
\right\}
=
0,
\qquad
\forall o_j\in B^*.
\end{equation}

Here, \(I\{\cdot\}\) denotes the indicator function, which equals 1 if the event is true and 0 otherwise. The function \(\mathrm{Collide}(o_j,o_0)\) indicates whether vehicle \(o_j\) physically collides with the ego vehicle \(o_0\). The first constraint states that at most one adversarial vehicle can collide with the ego vehicle, while the second constraint requires all support vehicles to remain non-colliding. Therefore, if an actual collision is triggered, the responsible vehicle must be the primary adversary.

In summary, the proposed problem formulation can be interpreted as follows. KG-ASG first uses collision knowledge to understand the current traffic scene and identify the target collision mode and key interacting vehicles. It then assigns primary-support roles to the selected participants and performs conditional adversarial trajectory generation under rule constraints. Finally, planner-controller closed-loop feedback is used to evaluate and refine the generated scenarios. This formulation provides a clear hierarchical structure for the proposed method and ensures that generated scenarios have not only adversarial effectiveness, but also explicit collision semantics, attribution structure, and executable boundaries.

\section{Methodology}
\begin{figure*}[t]
    \centering
    \includegraphics[width=\textwidth]{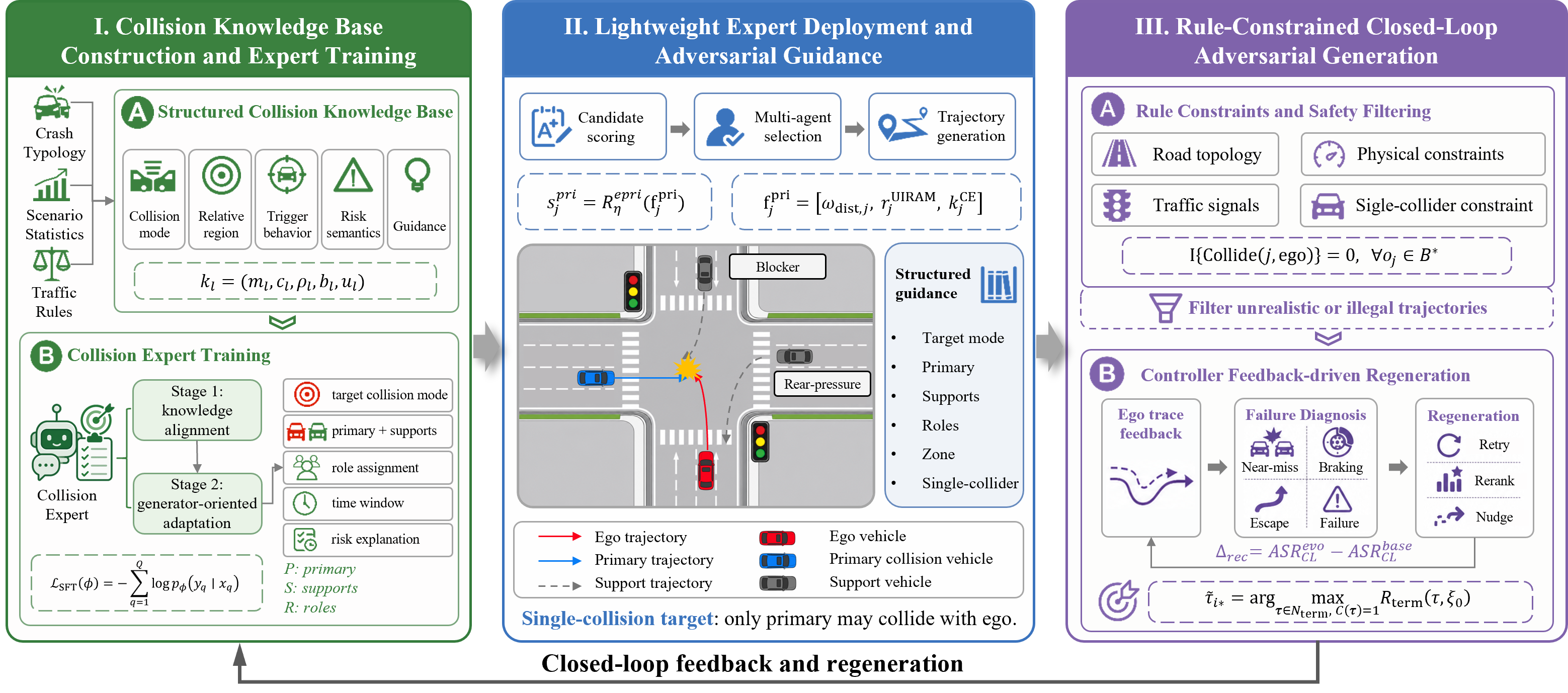}
    \caption{KG-ASG knowledge-guided closed-loop adversarial scenario generation framework. The framework uses high-level semantic priors to constrain low-level adversarial generation. The Collision Expert does not directly generate continuous trajectory points. Instead, it provides structured guidance, including the target collision mode, the primary adversary, support-vehicle roles, the conflict region, and the critical time window. Continuous trajectories are selected and optimized within the candidate trajectory space by the adversarial generation module. Rule constraints and planner-controller feedback are further incorporated to ensure executability, single-collider attribution, and closed-loop attack recovery.}
    \label{fig:kg_asg_framework}
\end{figure*}

Building on the above formulation, this paper proposes KG-ASG, a knowledge-guided closed-loop adversarial scenario generation framework. The central design of KG-ASG is to impose an attribution-preserving structure on adversarial generation rather than simply stacking additional optimization modules. KG-ASG first constructs a structured collision knowledge base and trains a lightweight Collision Expert to infer the target collision mode, the primary adversary, support-vehicle roles, and generation guidance. Based on the multimodal trajectory priors provided by a candidate-trajectory generator, KG-ASG performs primary-support vehicle selection and role-aware trajectory re-ranking, where the primary adversary is responsible for the final conflict and support vehicles are constrained to shape risk without becoming additional colliders. Finally, rule constraints and planner-controller feedback are incorporated to retry failed samples, re-rank candidates, and refine terminal trajectories, thereby generating high-risk, interpretable, attributable, and executable testing scenarios. The overall framework is shown in Fig.~\ref{fig:kg_asg_framework}.

To avoid notational ambiguity, \(S_i\) denotes the historical state sequence of vehicle \(o_i\), \(O(x)\) denotes the candidate vehicle set in scene \(x\), \(o_{i^*}\) denotes the unique primary adversary, \(B^*\) denotes the support vehicle set, and \(V^*=\{o_{i^*}\}\cup B^*\) denotes the final set of vehicles participating in adversarial generation. The trajectory-level rule constraint is denoted by \(C(\tau)\), while the scenario-level rule constraint is denoted by \(C(A)\). The support-role set is denoted by \(R_q\), and the structured generation guidance is denoted by \(u\).

\subsection{Collision Knowledge Base Construction and Expert Training}

The collision knowledge base provides retrievable, learnable, and interpretable accident-mechanism priors for adversarial generation. Following the NHTSA pre-crash typology, real-world pre-crash states, naturalistic driving statistics, traffic rules, and existing safety-critical scenario generation experience, different collision modes are organized into a unified structured representation~\cite{najm2007precrash}. Unlike a rule base that only records safe or unsafe labels, the proposed knowledge base explicitly associates pre-crash scene semantics, participant relations, triggering behaviors, and generation guidance. This design enables adversarial generation to move from low-level collision-proxy search toward targeted generation conditioned on collision semantics.

The \(l\)-th collision knowledge unit is defined as
\begin{equation}
k_l=
\left(
m_l,\,
c_l,\,
\rho_l,\,
b_l,\,
u_l
\right).
\end{equation}
Here, \(k_l\) denotes the \(l\)-th knowledge entry, \(m_l\) denotes the collision mode, such as rear-end conflict, crossing conflict, cut-in conflict, or merge conflict, \(c_l\) denotes the scene semantic condition associated with this collision mode, \(\rho_l\) denotes the key relative region, such as front, front-side, or crossing conflict zone, \(b_l\) denotes the typical triggering behavior, such as hard braking, late merging, aggressive crossing, or yielding failure, and \(u_l\) denotes the generation-guidance template for the adversarial generation module, including the target conflict region, critical time window, behavior priority, and trajectory-scoring bias.

Given a scene \(x\), knowledge retrieval aims to identify the collision mode that best matches the current scene semantics:
\begin{equation}
m^*
=
\arg\max_{m\in M_{\mathrm{col}}}
\mathrm{KBMatch}(x,m).
\end{equation}
Here, \(m^*\) denotes the target collision mode, \(M_{\mathrm{col}}\) denotes the set of candidate collision modes in the knowledge base, and \(\mathrm{KBMatch}(x,m)\) denotes the matching score between scene \(x\) and collision mode \(m\). This score jointly considers road topology, relative vehicle regions, motion trends, conflict time windows, and candidate-vehicle behavior patterns. The retrieved collision mode does not directly replace generator decisions. Instead, it provides a stable semantic anchor for the Collision Expert, allowing the model to align its reasoning with accident mechanisms.

Based on the knowledge base, this paper trains a lightweight Collision Expert. Let the expert model be denoted by \(f_{\phi}\), where \(\phi\) represents the model parameters. The training set is denoted by
\begin{equation}
D=
\left\{
(x_q,y_q)
\right\}_{q=1}^{Q},
\end{equation}
where \(x_q\) denotes the \(q\)-th training input, \(y_q\) denotes the corresponding structured expert output, and \(Q\) denotes the number of training samples. The expert output is defined as
\begin{equation}
y_q=
\left(
m_q^*,\,
i_q^*,\,
B_q^*,\,
R_q,\,
u_q
\right).
\end{equation}
Here, \(m_q^*\) denotes the target collision mode, \(i_q^*\) denotes the primary adversary, \(B_q^*\) denotes the support vehicle set, \(R_q\) denotes the support-role set, such as blocker or rear-pressure vehicle, and \(u_q\) denotes the conditional information required for subsequent generation, including the target region, critical time window, and generation preference.

The expert is trained using supervised fine-tuning:
\begin{equation}
L_{\mathrm{SFT}}(\phi)
=
-
\sum_{q=1}^{Q}
\log p_{\phi}(y_q\mid x_q).
\end{equation}
Here, \(L_{\mathrm{SFT}}\) denotes the supervised fine-tuning loss, and \(p_{\phi}(y_q\mid x_q)\) denotes the conditional probability that the Collision Expert with parameters \(\phi\) generates the target output \(y_q\) given the input \(x_q\). Through this objective, the model learns the mapping from traffic-scene semantics to collision modes, primary-support roles, and generation guidance.

During inference, the Collision Expert does not output continuous trajectory points. Instead, it decodes the scene state \(x\) and the candidate vehicle set \(O_K\) into structured collision semantics:
\begin{equation}
y=
\mathrm{Dec}_{\phi}
\left(
\mathrm{Prompt}(x,O_K)
\right).
\end{equation}
Here, \(\mathrm{Dec}_{\phi}(\cdot)\) denotes the decoding function of the expert model, \(\mathrm{Prompt}(x,O_K)\) denotes the structured input constructed from the scene, candidate vehicles, UIRAM risk scores, road context, and historical motion trends, and \(y\) denotes the generated structured guidance following the predefined output schema. The output is further processed by schema normalization and converted into the primary adversary, support roles, conflict region, time window, and trajectory re-ranking signals.

Table~\ref{tab:collision_expert_setup} summarizes the knowledge-base structure, training data, and key training settings of the Collision Expert. The expert only predicts the target collision mode, primary adversary, support roles, conflict time window, and structured generation guidance. Continuous trajectories are still selected, re-ranked, and filtered by the subsequent candidate-trajectory generation module. Although the SFT dataset is moderate in size, the expert is not trained to generate full trajectories. It only performs structured collision reasoning and role assignment over a constrained output schema, which substantially reduces the output complexity and improves sample efficiency.

\begin{table*}[t]
\centering
\caption{Collision Expert Training and Knowledge-Base Setup}
\label{tab:collision_expert_setup}
\footnotesize
\renewcommand{\arraystretch}{1.24}
\setlength{\tabcolsep}{6pt}
\begin{tabular}{p{0.17\textwidth}|p{0.76\textwidth}}
\toprule
Component & Summary \\
\midrule
Objective &
Two-stage LoRA supervised fine-tuning for structured collision reasoning. The expert predicts collision mode, primary adversary, support roles, conflict window, and generation guidance rather than continuous trajectories. \\

Training data &
Stage 1 uses 450 knowledge-alignment samples for collision-mode scoring, Top-K ranking, and generation guidance. Stage 2 uses 1050 generator-oriented samples with hard cases, primary-support assignment, and single-collider intent. \\

Knowledge schema &
Each entry contains collision mode, scene condition, relative conflict zone, trigger behavior, and generation guidance. \\

Backbone &
Mistral-7B-Instruct-v0.3 is used as the main expert. Qwen3-4B, Phi-4-mini, DeepSeek-R1-7B, and Granite-3.3-2B are evaluated as alternatives. \\

Training setup &
LoRA rank 16, alpha 32, and dropout 0.05 are used. Stage 1 uses learning rate \(3.0\times10^{-5}\) for 4 epochs. Stage 2 uses learning rate \(2.4\times10^{-5}\) for 3 epochs. \\

Inference output &
The scene-level prompt produces the target collision mode, primary opponent, support opponent, role assignment, conflict window, single-collider intent, and concise generation guidance. \\

Deployment &
The Stage-2 adapter is merged into the final Collision Expert and used for KG-ASG inference. \\
\bottomrule
\end{tabular}
\end{table*}

From a theoretical perspective, the collision knowledge base provides structured priors that decompose high-dimensional continuous trajectory search into a hierarchical problem of first determining collision semantics and then performing conditional generation. The Collision Expert acts as the semantic decision layer, enabling the generation process to be constrained by accident mechanisms and traffic-semantics consistency rather than relying solely on low-level collision proxies. The resulting structured output is then used as the conditional input for the next-stage adversarial generation module.

\subsection{Collision-Expert-Guided Primary-Support Adversarial Generation}

\begin{figure*}[t]
    \centering
    \includegraphics[width=0.8\textwidth]{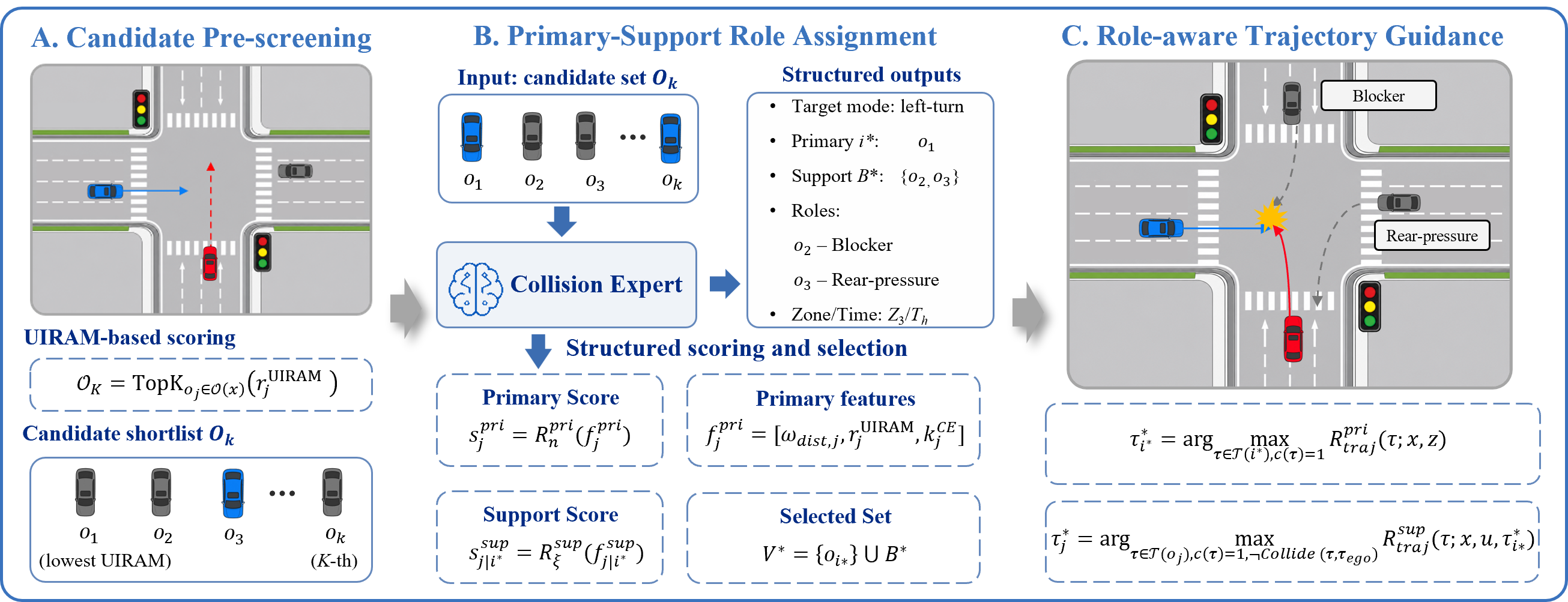}
    \caption{Role-aware primary-support adversarial generation framework. The Collision Expert provides structured semantic guidance, including the target collision mode, primary adversary, support vehicles, support roles, conflict region, and critical time window. Based on this guidance, KG-ASG performs UIRAM-based candidate pre-screening, primary-adversary selection, support-vehicle assignment, and role-aware trajectory re-ranking. The primary adversary is optimized to induce the main conflict, while support vehicles are constrained to shape the surrounding risk structure without becoming additional colliders.}
    \label{fig:primary_support_generation}
\end{figure*}

After expert training, the Collision Expert is integrated into the candidate-trajectory-based adversarial generation process. It provides the target collision mode, the primary adversary, support-vehicle roles, the conflict region, and the critical time window. Continuous trajectories are still selected and re-ranked within the candidate trajectory space. This design allows high-level collision semantics to constrain low-level trajectory generation while avoiding direct continuous-trajectory control by the language model, as shown in Fig.~\ref{fig:primary_support_generation}.

First, candidate vehicles are pre-screened using lightweight risk assessment. Let \(U(\cdot)\) denote the UIRAM risk assessment function, which jointly considers interaction relevance, intention uncertainty, collision probability, and collision consequence. The detailed computation is provided in~\cite{wang2026uiram}. For a candidate vehicle \(o_j\), its risk screening score is defined as
\begin{equation}
r_j^{\mathrm{UIRAM}}
=
U(o_j\mid x,o_0),
\qquad
o_j\in O(x).
\end{equation}
Here, \(r_j^{\mathrm{UIRAM}}\) denotes the UIRAM risk score of vehicle \(o_j\) with respect to the ego vehicle \(o_0\), \(x\) denotes the current traffic scene state, and \(O(x)\) denotes the candidate vehicle set. The Top-K candidate set is selected in descending order of \(r_j^{\mathrm{UIRAM}}\):
\begin{equation}
O_K
=
\mathrm{TopK}_{o_j\in O(x)}
\left(
r_j^{\mathrm{UIRAM}}
\right).
\end{equation}
Here, \(O_K\) denotes the candidate set sent to the Collision Expert for semantic reasoning. This step only reduces the search space. The final adversarial vehicle selection is still determined jointly by collision-knowledge reasoning and primary-support role modeling.

The primary adversary and support vehicles are then modeled separately. For a candidate vehicle \(o_j\), its suitability score for becoming the primary adversary is written as
\begin{equation}
s_j^{\mathrm{pri}}
=
R_{\eta}^{\mathrm{pri}}
\left(
f_j^{\mathrm{pri}}
\right),
\qquad
f_j^{\mathrm{pri}}
=
\left[
\omega_{\mathrm{dist},j},\,
r_j^{\mathrm{UIRAM}},\,
k_j^{\mathrm{CE}}
\right].
\end{equation}
Here, \(s_j^{\mathrm{pri}}\) denotes the primary-adversary suitability score of vehicle \(o_j\), \(R_{\eta}^{\mathrm{pri}}(\cdot)\) denotes the primary-adversary ranking function with parameter \(\eta\), \(f_j^{\mathrm{pri}}\) denotes the candidate feature vector, \(\omega_{\mathrm{dist},j}\) denotes the distance-risk score, \(r_j^{\mathrm{UIRAM}}\) denotes the UIRAM risk score, and \(k_j^{\mathrm{CE}}\) denotes the target-mode matching score provided by the Collision Expert. In this implementation, the parameters of \(R_{\eta}^{\mathrm{pri}}\) are calibrated on the validation set and kept fixed during all test experiments. When more training samples are available, this ranking function can also be replaced by a lightweight learnable ranker. No ranking parameter is tuned on the test set.

The primary adversary is selected by
\begin{equation}
o_{i^*}
=
\arg\max_{o_j\in O_K}
s_j^{\mathrm{pri}}.
\end{equation}

After determining the primary adversary, support vehicles are not selected according to the likelihood of colliding with the ego vehicle. Instead, they are conditionally evaluated by their ability to shape an effective risk structure around the primary adversary. For a candidate support vehicle \(o_j\), its conditional support score is defined as
\begin{equation}
s_{j\mid i^*}^{\mathrm{sup}}
=
R_{\zeta}^{\mathrm{sup}}
\left(
f_{j\mid i^*}^{\mathrm{sup}}
\right),
\end{equation}
\begin{equation}
f_{j\mid i^*}^{\mathrm{sup}}
=
\left[
k_{j\mid i^*}^{\mathrm{role}},\,
q_{j\mid i^*}^{\mathrm{geo}},\,
q_{j\mid i^*}^{\mathrm{time}},\,
-o_{j\mid i^*}^{\mathrm{overlap}}
\right].
\end{equation}
Here, \(s_{j\mid i^*}^{\mathrm{sup}}\) denotes the support suitability score of vehicle \(o_j\) conditioned on the primary adversary \(o_{i^*}\), \(R_{\zeta}^{\mathrm{sup}}(\cdot)\) denotes the support-vehicle ranking function, \(k_{j\mid i^*}^{\mathrm{role}}\) denotes the role-matching score, \(q_{j\mid i^*}^{\mathrm{geo}}\) denotes whether the vehicle is located in a position that can compress the ego vehicle's feasible avoidance space, \(q_{j\mid i^*}^{\mathrm{time}}\) denotes whether it can interfere near the primary conflict time window, and \(o_{j\mid i^*}^{\mathrm{overlap}}\) denotes its trajectory-corridor overlap with the primary adversary. This score encourages support vehicles to act as risk-structure shapers rather than additional colliders.

The support vehicle set is selected from the remaining candidates:
\begin{equation}
B^*
=
\mathrm{TopM}
\left(
s_{j\mid i^*}^{\mathrm{sup}}
\right),
\qquad
o_j\in O_K\setminus\{o_{i^*}\},
\qquad
|B^*|\leq 2.
\end{equation}
Here, \(B^*\) denotes the final support vehicle set. Motivated by pre-crash typology, typical multi-vehicle interaction complexity, and the cost of combinatorial search, KG-ASG keeps at most two support vehicles that satisfy role consistency, timing compatibility, non-collision constraints, and trajectory-corridor distinctiveness. The final adversarial vehicle set is
\begin{equation}
V^*
=
\{o_{i^*}\}\cup B^*.
\end{equation}

During trajectory generation, the primary adversary and support vehicles follow different scoring objectives. The primary adversary is responsible for inducing the main conflict, and its trajectory is selected by
\begin{equation}
\tau_{i^*}^{*}
=
\arg\max_{\tau\in T(o_{i^*}),\,C(\tau)=1}
R_{\mathrm{traj}}^{\mathrm{pri}}
\left(
\tau;x,u
\right).
\end{equation}
Here, \(\tau_{i^*}^{*}\) denotes the selected primary-adversary trajectory, \(T(o_{i^*})\) denotes the candidate trajectory space of the primary adversary, \(R_{\mathrm{traj}}^{\mathrm{pri}}\) denotes the primary trajectory ranking function, \(u\) denotes the target region, time window, and generation preference output by the Collision Expert, and \(C(\tau)=1\) indicates that the trajectory satisfies road, signal, dynamic, and interaction-safety constraints. The ranking function considers collision proxy, target-mode consistency, time-window matching, and target-region matching, while rule-related conditions are enforced as hard constraints.

Support vehicles are responsible for compressing the feasible avoidance space of the planner-controller, but they must not collide with the ego vehicle. For any \(o_j\in B^*\), the support trajectory is selected by
\begin{equation}
\tau_j^*
=
\arg\max_{\tau\in T(o_j),\,C(\tau)=1,\,\neg\mathrm{Collide}(\tau,\tau_0)}
R_{\mathrm{traj}}^{\mathrm{sup}}
\left(
\tau;x,u,\tau_{i^*}^{*}
\right).
\end{equation}
Here, \(\tau_j^*\) denotes the selected trajectory of support vehicle \(o_j\), \(R_{\mathrm{traj}}^{\mathrm{sup}}\) denotes the support trajectory ranking function, \(\tau_0\) denotes the ego reference trajectory, and \(\neg\mathrm{Collide}(\tau,\tau_0)\) means that the support trajectory must not collide with the ego vehicle. The support trajectory ranking function mainly evaluates non-collision pressure, role consistency, timing compatibility, naturalness, and corridor distinctiveness from the primary adversary. In this implementation, the overlap penalty between support vehicles and the primary adversary uses validation-fixed parameters, with a default overlap penalty coefficient of \(0.35\) and a corridor margin of \(0.75\,\mathrm{m}\), and these parameters remain fixed during testing.

Thus, the primary adversary induces the main conflict, while support vehicles compress the feasible avoidance space without colliding with the ego vehicle. This design transforms multi-vehicle adversarial generation from multi-vehicle collision seeking into a conditional generation problem in which a single primary adversary is responsible for the main conflict and support vehicles shape the risk structure. It improves scenario interpretability and reduces abnormal multi-vehicle collision artifacts.

\subsection{Rule-Constrained and  Closed-Loop Regeneration}

Although the Collision Expert and adversarial generation module can produce high-risk candidate scenarios, unconstrained generation may obtain false attack success through lane violations, abnormal acceleration, off-road driving, or simultaneous multi-vehicle collisions. Therefore, KG-ASG introduces a rule checker that integrates road topology, traffic signals, dynamic limits, environmental-vehicle safety, and single-collider constraints into candidate filtering and closed-loop regeneration.

For any vehicle trajectory \(\tau\), the rule constraint function is defined as
\begin{equation}
C(\tau)
=
C_{\mathrm{road}}(\tau)
\wedge
C_{\mathrm{signal}}(\tau)
\wedge
C_{\mathrm{dyn}}(\tau)
\wedge
C_{\mathrm{int}}(\tau).
\end{equation}
Here, \(C(\tau)\in\{0,1\}\) denotes whether the trajectory satisfies all constraints, \(C_{\mathrm{road}}\) denotes road-topology and drivable-area constraints, \(C_{\mathrm{signal}}\) denotes traffic-signal and right-of-way constraints, \(C_{\mathrm{dyn}}\) denotes dynamic constraints such as velocity, acceleration, lateral acceleration, and jerk, and \(C_{\mathrm{int}}\) denotes inter-vehicle interaction-safety constraints. For a complete adversarial scenario \(A\), the scenario-level constraint is
\begin{equation}
C(A)
=
\bigwedge_{(o_j,\tau_j)\in A}
C(\tau_j).
\end{equation}
If \(C(A)=0\), the candidate scenario is not accepted as the final output. Thus, rule compliance is not treated as a soft weighted penalty, but as a hard gate for candidate filtering.

To ensure clear collision attribution, KG-ASG further imposes the single-collider constraint:
\begin{equation}
\sum_{o_j\in V^*}
I
\left\{
\mathrm{Collide}(o_j,o_0)
\right\}
\leq 1,
\end{equation}
\begin{equation}
I
\left\{
\mathrm{Collide}(o_j,o_0)
\right\}
=
0,
\qquad
\forall o_j\in B^*.
\end{equation}
The first constraint states that at most one adversarial vehicle can collide with the ego vehicle \(o_0\). The second constraint requires all support vehicles to remain non-colliding. Therefore, if a collision occurs, the colliding vehicle must be the primary adversary \(o_{i^*}\). This design provides explicit collision attribution and avoids explanation failure caused by uncontrolled multi-vehicle impacts. \par

The closed-loop stage treats the planner-controller \(\pi_c\) strictly as the system under test. It does not update planner-controller parameters or train an ego policy. Instead, KG-ASG uses the executed ego trajectory, collision outcome, minimum distance, and failure type returned by the planner-controller to retry, re-rank, and terminally refine adversarial candidates. Therefore, the proposed closed-loop optimization is rollout-level feedback regeneration rather than controller training. 

After rule filtering, the generated scenario is executed by the planner-controller \(\pi_c\). The adversarial scenario generated at round \(r\) is defined as
\begin{equation}
A^{(r)}
=
G_{\theta}
\left(
x;u,\tau_0^{(r)},p^{(r)}
\right),
\end{equation}
where \(A^{(r)}\) denotes the adversarial scenario at round \(r\), \(G_{\theta}\) denotes the adversarial generation module with parameter \(\theta\), \(u\) denotes the structured generation guidance, \(\tau_0^{(r)}\) denotes the ego reference trajectory used at round \(r\), and \(p^{(r)}\) denotes the retry profile, such as timing synchronization, steering trap, brake-delay pressure, or terminal refinement.

After execution, the closed-loop feedback is obtained as
\begin{equation}
h^{(r)}
=
H
\left(
x,A^{(r)};\pi_c
\right)
=
\left(
\xi_0^{(r)},\,
I_{\mathrm{col}}^{(r)},\,
d_{\min}^{(r)},\,
e^{(r)}
\right).
\end{equation}
Here, \(h^{(r)}\) denotes the closed-loop feedback at round \(r\), \(H(\cdot)\) denotes the closed-loop execution function, \(\xi_0^{(r)}\) denotes the ego trajectory executed by the planner-controller, \(I_{\mathrm{col}}^{(r)}\) indicates whether the ego vehicle collides, \(d_{\min}^{(r)}\) denotes the minimum distance, and \(e^{(r)}\) denotes the failure or escape type. If \(I_{\mathrm{col}}^{(r)}=1\), the closed-loop process stops. If \(I_{\mathrm{col}}^{(r)}=0\), the executed ego trajectory is used as the next-round attack reference, and the retry profile is selected according to the failure type:
\begin{equation}
\tau_0^{(r+1)}
=
\xi_0^{(r)},
\qquad
p^{(r+1)}
=
P(e^{(r)}).
\end{equation}
Here, \(P(\cdot)\) denotes the mapping from failure type to retry profile. Early-braking escape delays risk exposure and compresses the remaining braking margin. Steering escape strengthens lateral blocking. Timing mismatch re-aligns the arrival times of the primary adversary and the ego vehicle at the conflict region.

In the final closed-loop refinement round, KG-ASG applies terminal refinement to samples that have formed near misses but are still avoided by the planner-controller. Given the current primary trajectory \(\tau_{i^*}\) and the feedback ego trajectory \(\xi_0\), the terminal refinement candidate set is constructed as
\begin{equation}
\begin{aligned}
N_{\mathrm{term}}\!\left(\tau_{i^*}, \xi_0\right)
= {} & \{\tau_{i^*}\}
\cup \mathrm{Nudge}\!\left(\tau_{i^*}, \xi_0\right) \\
& \cup \mathrm{ResampleNudge}\!\left(\tau_{i^*}, \xi_0\right) .
\end{aligned}
\label{eq:nterm}
\end{equation}
Here, \(N_{\mathrm{term}}\) denotes the terminal refinement candidate set, \(\mathrm{Nudge}(\cdot)\) denotes the candidate set obtained by small legal terminal modifications to the current trajectory, and \(\mathrm{ResampleNudge}(\cdot)\) denotes the candidate set obtained by resampling terminal perturbations around the feedback ego trajectory. The refined primary trajectory is selected from feasible candidates:
\begin{equation}
\tilde{\tau}_{i^*}
=
\arg\max_{\tau\in N_{\mathrm{term}},\,C(\tau)=1}
R_{\mathrm{term}}
\left(
\tau,\xi_0
\right).
\end{equation}
Here, \(\tilde{\tau}_{i^*}\) denotes the terminal-refined primary trajectory, and \(R_{\mathrm{term}}\) denotes the terminal refinement ranking function. It evaluates collision proxy, minimum-distance closeness, escape-path pressure, near-distance duration, and consistency with the target conflict time window within the feasible candidate set. Since ranking is performed only among feasible candidates, terminal refinement cannot increase attack strength by violating traffic rules or dynamic feasibility.

The closed-loop adversarial generation process is summarized as
\begin{equation}
A^{(0)}
\rightarrow
h^{(0)}
\rightarrow
A^{(1)}
\rightarrow
h^{(1)}
\rightarrow
\cdots
\rightarrow
A^{(R_{\max})}.
\end{equation}
Here, \(R_{\max}\) denotes the maximum number of closed-loop regeneration rounds, which is set to 5 in this paper. The process does not update planner-controller parameters. Instead, it uses the executed ego trajectory, failure type, and minimum distance returned by the planner-controller to retry, re-rank, and terminally refine adversarial generation.

Finally, closed-loop recovery gain is used to evaluate the effectiveness of feedback-driven regeneration:
\begin{equation}
\Delta_{\mathrm{rec}}(\pi_c)
=
\mathrm{ASR}_{\mathrm{CL}}^{\mathrm{evo}}(\pi_c)
-
\mathrm{ASR}_{\mathrm{CL}}^{\mathrm{base}}(\pi_c).
\end{equation}
Here, \(\Delta_{\mathrm{rec}}(\pi_c)\) denotes the closed-loop recovery gain for planner-controller \(\pi_c\), \(\mathrm{ASR}_{\mathrm{CL}}^{\mathrm{evo}}(\pi_c)\) denotes the closed-loop attack success rate after enabling regeneration, and \(\mathrm{ASR}_{\mathrm{CL}}^{\mathrm{base}}(\pi_c)\) denotes the closed-loop baseline attack success rate without regeneration. This metric avoids the bias of directly evaluating closed-loop scenarios using open-loop attack rates, since a reactive planner-controller can actively brake, avoid, or re-plan. The complete procedure is summarized in Algorithm~\ref{alg:closed_loop_regeneration}.

\begin{algorithm}[t]
\caption{Controller-conditioned closed-loop regeneration}
\label{alg:closed_loop_regeneration}
\footnotesize
\begin{algorithmic}[1]
\REQUIRE traffic scene \(x\); planner-controller \(\pi_c\); Collision Expert \(f_{\phi}\); adversarial generator \(G_{\theta}\); maximum retry rounds \(R_{\max}\).
\ENSURE Final adversarial scenario \(A^*\).

\STATE Obtain the candidate shortlist \(O_K\) by UIRAM-based risk pre-screening.

\STATE Query the Collision Expert and obtain structured guidance
\[
g=f_{\phi}(x,O_K,K)=(m^*,o_{i^*},B^*,R_q,u),
\qquad
V^*=\{o_{i^*}\}\cup B^*.
\]

\STATE Initialize \(\tau_0^{(0)}\) from replay or prediction, and set \(A^*\leftarrow\emptyset\), \(J^*\leftarrow-\infty\).

\FOR{\(r=0,\ldots,R_{\max}\)}

\STATE Generate a candidate adversarial scenario
\[
A^{(r)}
=
G_{\theta}(x;u,\tau_0^{(r)},p^{(r)}).
\]

\STATE Apply rule, physical, single-collider, and support no-collision gates. If \(C(A^{(r)})=0\), reject the candidate and continue.

\STATE Execute \(A^{(r)}\) with controller \(\pi_c\) and obtain feedback
\[
h^{(r)}
=
H(x,A^{(r)};\pi_c)
=
(\xi_0^{(r)},I_{\mathrm{col}}^{(r)},d_{\min}^{(r)},e^{(r)}).
\]

\STATE Compute the closed-loop score \(J_{\mathrm{CL}}(A^{(r)})\) using collision outcome, minimum distance, rule compliance, and trajectory naturalness.

\IF{\(J_{\mathrm{CL}}(A^{(r)})>J^*\)}
    \STATE Update
    \[
    A^*\leftarrow A^{(r)},
    \qquad
    J^*\leftarrow J_{\mathrm{CL}}(A^{(r)}).
    \]
\ENDIF

\IF{\(I_{\mathrm{col}}^{(r)}=1\)}
    \STATE Terminate the loop and return \(A^*\).
\ENDIF

\STATE Otherwise, update the ego reference and retry profile
\[
\tau_0^{(r+1)}
=
\xi_0^{(r)},
\qquad
p^{(r+1)}
=
P(e^{(r)}).
\]

\STATE The retry profile is selected from timing synchronization, steering trap, near-miss push, or brake-delay pressure.

\IF{\(r = R_{\max}\)}
    \STATE Apply terminal refinement to the primary trajectory using feasible candidates defined in (34)--(35).
\ENDIF

\ENDFOR

\STATE Return \(A^*\).

\end{algorithmic}
\end{algorithm}

\section{Experiments}

We systematically evaluate KG-ASG on WOMD scenarios reconstructed in MetaDrive~\cite{ettinger2021womd,li2022metadrive}. Each scenario lasts 9 s, where the first 1 s is used as historical observation and the remaining 8 s is used for adversarial trajectory generation. To ensure comparability with existing safety-critical scenario generation studies, we follow a candidate-trajectory-based adversarial generation protocol. The trajectory proposal backbone is DenseTNT or an equivalent candidate trajectory generator~\cite{gu2021densetnt}. The number of opponent trajectory candidates is set to \(M=32\). Unless otherwise specified, the ego rollout buffer is set to \(N=5\), and the discount factor is set to \(\alpha=0.99\). KG-ASG builds on the candidate-trajectory adversarial generation paradigm and augments it with a collision knowledge base, a Collision Expert, a primary-support adversarial generation mechanism, and a rule-constrained closed-loop regeneration module. Under our hardware setting with dual Intel Xeon Platinum 8375C CPUs, 503 GiB RAM, and 8 NVIDIA RTX 4090 GPUs, KG-ASG remains efficient, requiring 5.53 s per scene without retry and 10.91 s per scene with up to five closed-loop retry rounds. \par

The experiments are organized as a progressive evidence chain. We first describe the evaluation protocols, metrics, and knowledge-base statistics to clarify the experimental basis of KG-ASG. We then evaluate open-loop adversarial performance under Replay and RL ego policies to examine whether KG-ASG preserves strong attack capability. Next, we analyze rule and physical constraints to verify whether high attack success is achieved without relying on invalid map violations or physically implausible trajectories. We further conduct expert and primary-support ablations to test whether KG-ASG improves collision attribution rather than merely increasing collision frequency. Finally, we evaluate closed-loop feedback regeneration against reactive planner-controllers. \par

For a fair comparison at the protocol-level, we follow the evaluation setting WOMD/MetaDrive and the metric definitions used in SAGE~\cite{nie2026sage}. KG-ASG is implemented in the same candidate-trajectory generation environment, using the same scenario split, trajectory budget, ego-policy setting, and evaluation scripts as the SAGE protocol. For cross-method comparison, non-KG-ASG baselines in Tables~\ref{tab_replay_benchmark} and~\ref{tab_rl_benchmark} are taken from the corresponding papers under the same or comparable WOMD/MetaDrive protocol when available. Since random seeds and some filtering details of the reported baselines are not fully accessible, these results from the cross-methods are interpreted as protocol-level comparisons rather than strict implementation-level rankings. In contrast, KG-ASG, all KG-ASG variants, no-expert comparisons, and closed-loop recovery experiments are reproduced in our implementation environment under the same scenario split, trajectory budget, and metric definitions. Therefore, the causal conclusions of this paper are drawn primarily from same-environment KG-ASG ablations.

\subsection{Experimental Setup and Metrics}

We adopt two complementary evaluation protocols. We follow a candidate-trajectory-based adversarial generation setting, where KG-ASG is implemented on top of multimodal opponent trajectory proposals and evaluated under Replay and learned ego policies. Replay indicates that the ego vehicle follows the logged or predicted trajectory, IDM denotes a rule-based reactive car-following controller, and Pretrained denotes a learned ego policy. The second is a SAGE-style open-loop quality protocol, which compares Rule, CAT, KING~\cite{hanselmann2022king}, AdvTrajOpt~\cite{zhang2022advtrajopt}, SEAL~\cite{stoler2025seal}, GOOSE~\cite{ransiek2024goose}, SAGE~\cite{nie2026sage}, and KG-ASG in terms of attack strength, kinematic plausibility, map compliance, and distributional distance. For SAGE, we report the strongest adversarial setting \(w_{\mathrm{adv}}=1.0\). \par

The main evaluation metrics include attack success rate, reward-based metrics, penalty-based metrics, distributional distance metrics, and closed-loop recovery metrics. ASR or Attack Succ. denotes the proportion of generated scenarios that trigger the target collision or safety-critical event under a given ego policy or planner-controller. It is computed as
\begin{equation}
\mathrm{ASR}
=
\frac{1}{N_{\mathrm{test}}}
\sum_{n=1}^{N_{\mathrm{test}}}
I_{\mathrm{col}}^n .
\end{equation}
Here, \(N_{\mathrm{test}}\) denotes the number of test scenarios, and \(I_{\mathrm{col}}^n\) indicates whether the \(n\)-th scenario triggers a target collision or safety-critical event. Adv. Reward denotes the adversarial reward defined by the benchmark, where a larger value indicates stronger attack intensity. Kine. Pen. measures the kinematic plausibility penalty. Crash Obj. and Cross Line measure irrelevant-object collision and line-crossing or map-violation penalties. WD-Accel, WD-Vel., and WD-Yaw denote the Wasserstein distances between generated trajectories and naturalistic driving data in acceleration, velocity, and yaw distributions. These reward and penalty terms follow the public metric definitions used in the comparison benchmark~\cite{nie2026sage}. Therefore, this paper focuses on relative trends under the same metric protocol rather than redefining internal weights for each submetric. For closed-loop evaluation, the baseline and KG-ASG regeneration results are compared under the same planner-controller and the same scenario set. Recovery Gain is defined as
\begin{equation}
\mathrm{Recovery\ Gain}
=
\mathrm{CL\text{-}ASR}_{\mathrm{KG\text{-}ASG}}
-
\mathrm{CL\text{-}ASR}_{\mathrm{base}} .
\end{equation}

\begin{figure*}[t]
\centering
\includegraphics[width=0.8\textwidth]{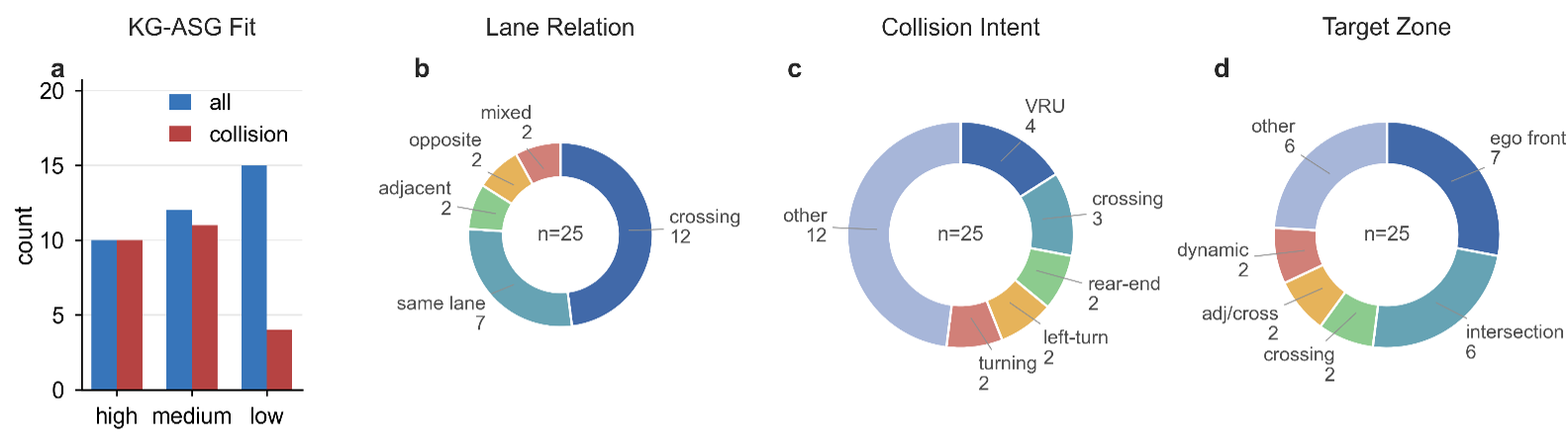}
\caption{Statistics of the constructed collision knowledge base. 
(a) KG-ASG fit distribution for all knowledge entries and collision-related entries. 
(b) Distribution of lane relations in collision-related scenarios. 
(c) Distribution of collision intent categories. 
(d) Distribution of target conflict zones. 
The results show that the knowledge base is generation-oriented and covers diverse spatial relations, collision semantics, and target-region guidance, which provides structured priors for Collision Expert training and downstream adversarial scenario generation.}
\label{fig:kb_overview}
\end{figure*}

Fig.~\ref{fig:kb_overview} summarizes the collision knowledge base used for Collision Expert training. Collision-related entries are mainly concentrated in high and medium KG-ASG-fit levels, indicating that most entries can be converted into actionable generation guidance rather than remaining as abstract traffic rules. The lane-relation, collision-intent, and target-zone distributions further show that the knowledge base covers diverse interaction patterns, including crossing, same-lane, adjacent, opposite, VRU, rear-end, turning, ego-front, intersection, and dynamic conflict cases. These statistics support the use of the knowledge base as a generation-oriented semantic prior for downstream adversarial scenario generation.

\subsection{Open-Loop Adversarial Performance and Executability}

This section evaluates open-loop adversarial effectiveness and executability. We first compare KG-ASG with representative baselines under the Replay policy to examine whether knowledge-guided primary-support generation preserves strong attack capability. We then provide a qualitative comparison to show how KG-ASG differs from direct trajectory perturbation. The RL-policy benchmark further tests whether the generated attacks remain effective against a learned ego policy. Finally, the factorized constraint ablation examines whether high attack success relies on invalid map violations or physically implausible trajectories. \par

\begin{table*}[t]
\centering
\caption{Open-Loop Generation Benchmark Against Replay Policy}
\label{tab_replay_benchmark}
\footnotesize
\renewcommand{\arraystretch}{1.20}
\setlength{\tabcolsep}{4.0pt}
\begin{tabular}{l|cc|ccc|ccc}
\toprule
Method & Attack Succ. \(\uparrow\) & Adv. Reward \(\uparrow\) & Kine. Pen. \(\downarrow\) & Crash Obj. \(\downarrow\) & Cross Line \(\downarrow\) & WD-Accel \(\downarrow\) & WD-Vel. \(\downarrow\) & WD-Yaw \(\downarrow\) \\
\midrule
Rule & \textbf{100.00\%} & \textbf{5.048} & 5.614 & 1.734 & 7.724 & 2.080 & 8.546 & 0.204 \\
CAT & 91.35\% & 3.961 & 3.143 & 2.466 & 9.078 & \underline{1.556} & \underline{7.233} & 0.225 \\
KING & 40.85\% & 2.243 & 3.434 & 3.126 & 6.056 & \textbf{0.972} & 255.5 & \textbf{0.098} \\
AdvTrajOpt & 70.46\% & 2.652 & \underline{2.775} & 2.547 & 10.16 & 1.754 & 6.177 & 0.268 \\
SEAL & 63.93\% & 1.269 & \textbf{2.423} & 2.732 & 11.612 & 1.544 & 6.959 & 0.202 \\
GOOSE & 36.07\% & 2.378 & 21.32 & 4.426 & 14.48 & 6.368 & 8.286 & \underline{0.154} \\
SAGE & 76.15\% & \underline{4.121} & 2.479 & \underline{0.731} & \underline{1.084} & 2.098 & 8.088 & 0.184 \\
KG-ASG & \underline{92.60\%} & 3.457 & 3.018 & \textbf{0.020} & \textbf{0.000} & 2.223 & \textbf{0.886} & 0.233 \\
\bottomrule
\end{tabular}
\end{table*}

As shown in Table~\ref{tab_replay_benchmark}, KG-ASG achieves an Attack Succ. of 92.60\% under the Replay policy, substantially outperforming SAGE at 76.15\% and approaching the strong CAT-style attack baseline. More importantly, KG-ASG obtains a Crash Obj. of only 0.020 and a Cross Line value of 0.000, which are significantly lower than those of other baselines. This indicates that KG-ASG does not achieve high attack success by leaving the map, crossing lines, or colliding with irrelevant objects. Instead, it concentrates the attack on valid ego-adversary interactions. KG-ASG is not the best on every realism-related metric, and its Kine. Pen. is higher than that of SAGE. Its advantage lies in a more safety-testing-oriented balance among high attack success, extremely low irrelevant-object collision, zero line-crossing violation, and a velocity distribution closer to naturalistic driving. Therefore, KG-ASG is not intended to dominate all distributional realism metrics, but to achieve a safety-testing-oriented trade-off among attack success, map compliance, collision attribution, and trajectory plausibility.

\begin{figure}[t]
    \centering
    \includegraphics[width=\columnwidth]{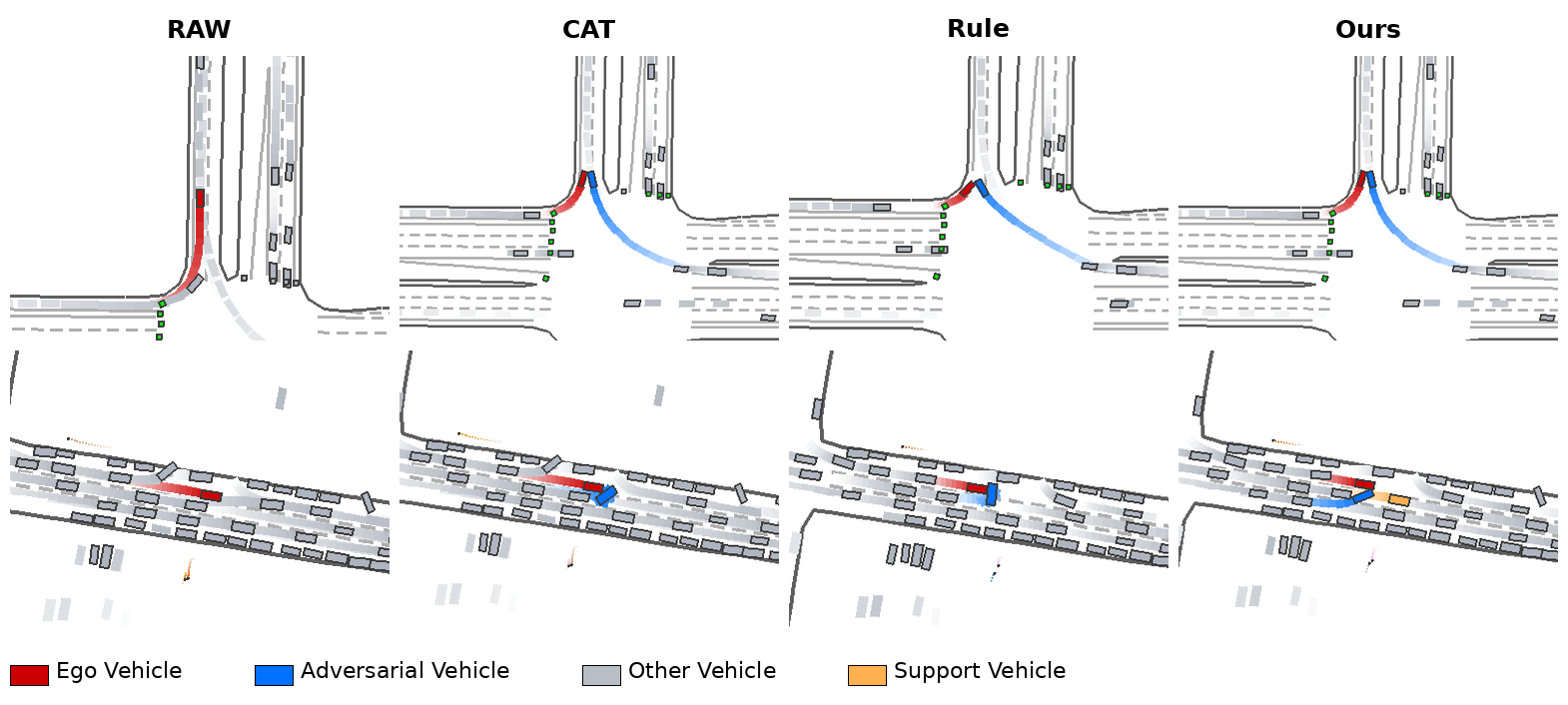}
    \caption{Method-level qualitative comparison. KG-ASG preserves the original traffic context while organizing selected vehicles into a primary adversary and support or pressure vehicles. The generated risk structure has explicit collision attribution, where the primary adversary is responsible for the final conflict and support vehicles compress the feasible avoidance space of the planner-controller.}
    \label{fig:method_comparison}
\end{figure}

Fig.~\ref{fig:method_comparison} provides a qualitative counterpart to the Replay-policy comparison. RAW mainly reproduces logged interactions, whereas CAT and Rule amplify risks through direct trajectory perturbation. KG-ASG instead preserves the original traffic context and organizes existing vehicles into a primary-support structure. The primary adversary is responsible for the final conflict, while the support vehicle shapes the local risk structure without becoming another collider. This observation supports the quantitative finding that KG-ASG maintains strong attack performance while providing clearer collision attribution.

\begin{table*}[t]
\centering
\caption{Open-Loop Generation Benchmark Against RL Policy}
\label{tab_rl_benchmark}
\footnotesize
\renewcommand{\arraystretch}{1.20}
\setlength{\tabcolsep}{4.0pt}
\begin{tabular}{l|cc|ccc|ccc}
\toprule
Method & Attack Succ. \(\uparrow\) & Adv. Reward \(\uparrow\) & Kine. Pen. \(\downarrow\) & Crash Obj. \(\downarrow\) & Cross Line \(\downarrow\) & WD-Accel \(\downarrow\) & WD-Vel. \(\downarrow\) & WD-Yaw \(\downarrow\) \\
\midrule
Rule & \textbf{65.57\%} & \textbf{2.761} & 113.7 & 1.803 & 6.148 & 10.85 & 13.47 & 0.336 \\
CAT & 30.33\% & 1.319 & 3.039 & 2.623 & 6.967 & \underline{1.539} & 8.877 & \underline{0.187} \\
KING & 19.14\% & 1.148 & \underline{2.596} & 3.114 & 5.857 & \textbf{0.983} & 259.1 & \textbf{0.097} \\
AdvTrajOpt & 19.40\% & 0.992 & 2.779 & 2.459 & 9.973 & 1.749 & 6.187 & 0.269 \\
SEAL & 31.40\% & 0.752 & 2.684 & 3.030 & 11.98 & 1.563 & 8.267 & 0.267 \\
GOOSE & 12.46\% & 0.667 & 15.47 & 3.507 & 11.45 & 4.662 & 8.070 & 0.152 \\
SAGE & 28.42\% & 1.400 & \textbf{2.496} & \underline{0.792} & \underline{1.366} & 2.098 & 8.114 & 0.188 \\
KG-ASG & \underline{39.00\%} & \underline{1.870} & 3.002 & \textbf{0.020} & \textbf{0.000} & 2.217 & \textbf{0.838} & 0.202 \\
\bottomrule
\end{tabular}
\end{table*}

As shown in Table~\ref{tab_rl_benchmark}, KG-ASG achieves an Attack Succ. of 39.00\% under the RL policy, outperforming SAGE at 28.42\% and CAT at 30.33\%. This indicates that expert-guided primary-support attacks remain effective even when the ego behavior is generated by a learned policy rather than replayed trajectories. Meanwhile, KG-ASG keeps Crash Obj. and Cross Line near zero, showing that the improvement does not come from invalid map violations or irrelevant-object collisions. Similar to the Replay-policy results, KG-ASG does not achieve the lowest Kine. Pen., but it provides a stronger balance among attack effectiveness, map compliance, and valid collision attribution.

\begin{table*}[t]
\centering
\caption{Factorized Ablation of Rule and Physical Constraints}
\label{tab_factorized_constraints}
\footnotesize
\renewcommand{\arraystretch}{1.20}
\setlength{\tabcolsep}{4.0pt}
\begin{tabular}{l|cc|cccccc}
\toprule
Setting & Rule & Physical & ASR \(\uparrow\) & Crash Obj. \(\downarrow\) & Accel Viol. \(\downarrow\) & Lat. Accel Viol. \(\downarrow\) & Road Dep. \(\downarrow\) & Signal Viol. \(\downarrow\) \\
\midrule
No constraints & \xmark & \xmark & \textbf{96.20\%} & \underline{1.720} & \underline{58.80\%} & 25.40\% & 7.40\% & 3.60\% \\
Rule only & \cmark & \xmark & \underline{92.60\%} & \textbf{0.020} & 89.80\% & \underline{20.00\%} & 0.80\% & \textbf{0.00\%} \\
Physical only & \xmark & \cmark & 84.40\% & \underline{1.720} & \textbf{13.00\%} & \textbf{19.40\%} & \underline{0.20\%} & \underline{0.90\%} \\
Rule + Physical & \cmark & \cmark & 92.40\% & \textbf{0.020} & \textbf{13.00\%} & \textbf{19.40\%} & \underline{0.20\%} & \textbf{0.00\%} \\
\bottomrule
\end{tabular}
\end{table*}

Table~\ref{tab_factorized_constraints} further explains how executability is maintained. Rule only substantially reduces irrelevant-object collision and traffic-rule violations, but it does not smooth trajectories and therefore can retain aggressive yet rule-compliant candidates with high acceleration violations. Physical only reduces acceleration violations, but it cannot remove irrelevant-object collisions. Combining rule and physical constraints maintains a high ASR of 92.40\% while achieving low crash-object, acceleration, road-departure, and signal-violation rates. This confirms that rule constraints mainly ensure semantic and traffic legality, whereas physical constraints improve kinematic executability.

\subsection{Collision Expert and Primary-Support Attribution}

This section examines whether the Collision Expert and the primary-support mechanism improve attack attribution rather than merely increasing collision frequency. All variants use the same WOMD and MetaDrive scenarios, the same candidate trajectory generator, and the same trajectory budget \(M=32\). Stage1 introduces structured collision reasoning and primary-support role assignment. Stage2 further injects role-aware objectives into trajectory generation. KG-ASG Full adds support no-collision, primary-only collision, fallback promotion, and rule constraints.

The Collision Expert is evaluated from two perspectives. The first benchmark evaluates structured collision reasoning over the collision knowledge base, including collision mode, intent, and structured output fields. The second benchmark evaluates target collision-mode and key-actor recognition on Waymo scenario fragments. Stage1 mainly performs knowledge-alignment training to stabilize schema-compliant structured output. Stage2 further adapts the model to the generation task, making its outputs more suitable for downstream primary-support assignment and trajectory guidance. Evaluation metrics include top-1 scenario, top-1 intent, and top-3 hit, which measure output-format stability and collision-semantic recognition.

\begin{figure}[t]
    \centering
    \includegraphics[width=0.8\columnwidth]{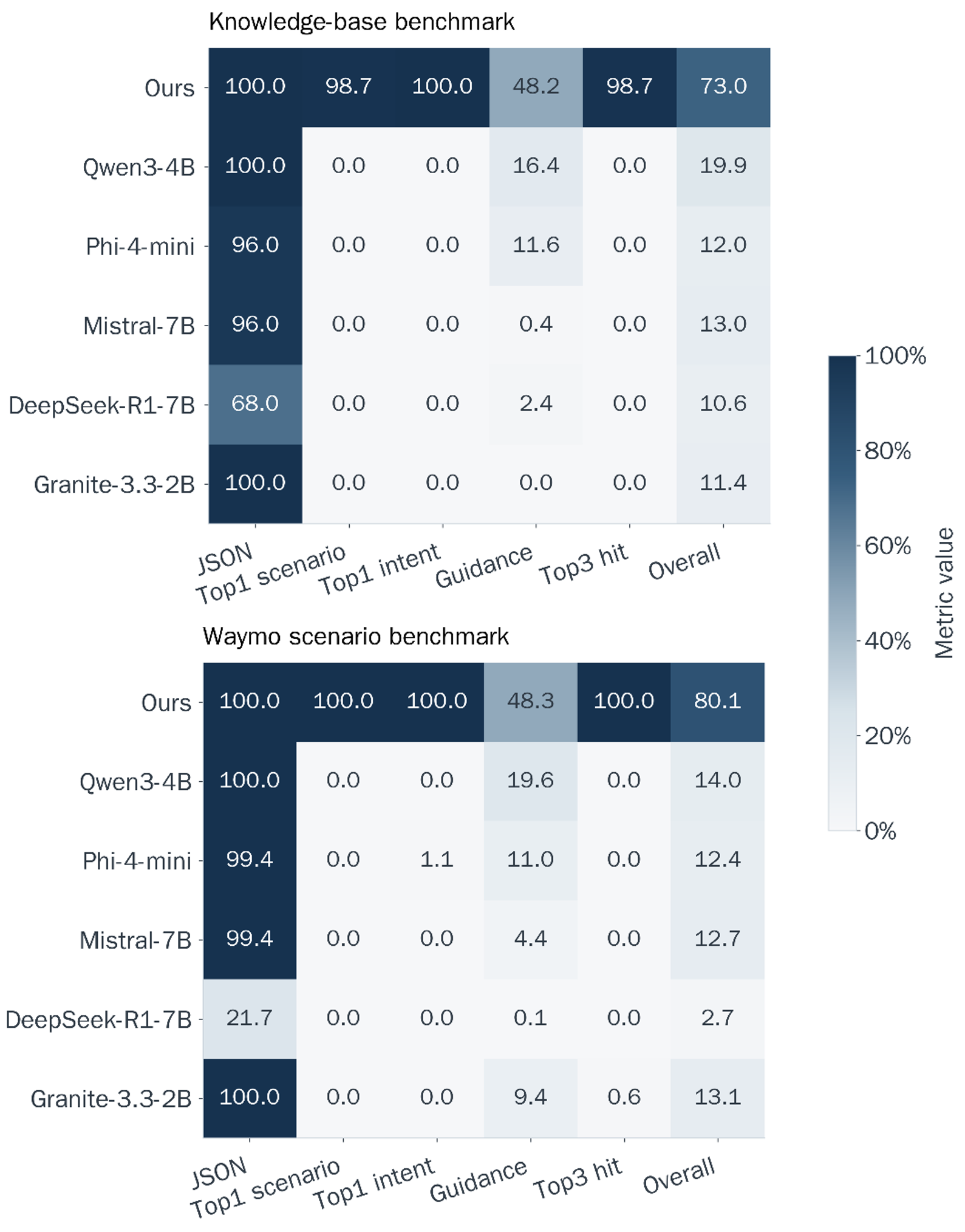}
    \caption{Stage-wise improvement of the Collision Expert. The staged training process improves both structured knowledge alignment and scenario-level collision reasoning, making the expert more reliable for downstream primary-support assignment.}
    \label{fig_stagewise_expert}
\end{figure}

As shown in Fig.~\ref{fig_stagewise_expert}, staged training improves the stability of the Collision Expert on both structured knowledge reasoning and scenario-level collision recognition. Stage1 mainly improves knowledge alignment and schema-compliant outputs, while Stage2 further enhances collision-mode recognition, intent inference, and generation-guidance quality. This suggests that generation-oriented adaptation is necessary for converting collision knowledge into reliable primary-support guidance.

\begin{figure}[t]
    \centering
    \includegraphics[width=0.75\columnwidth]{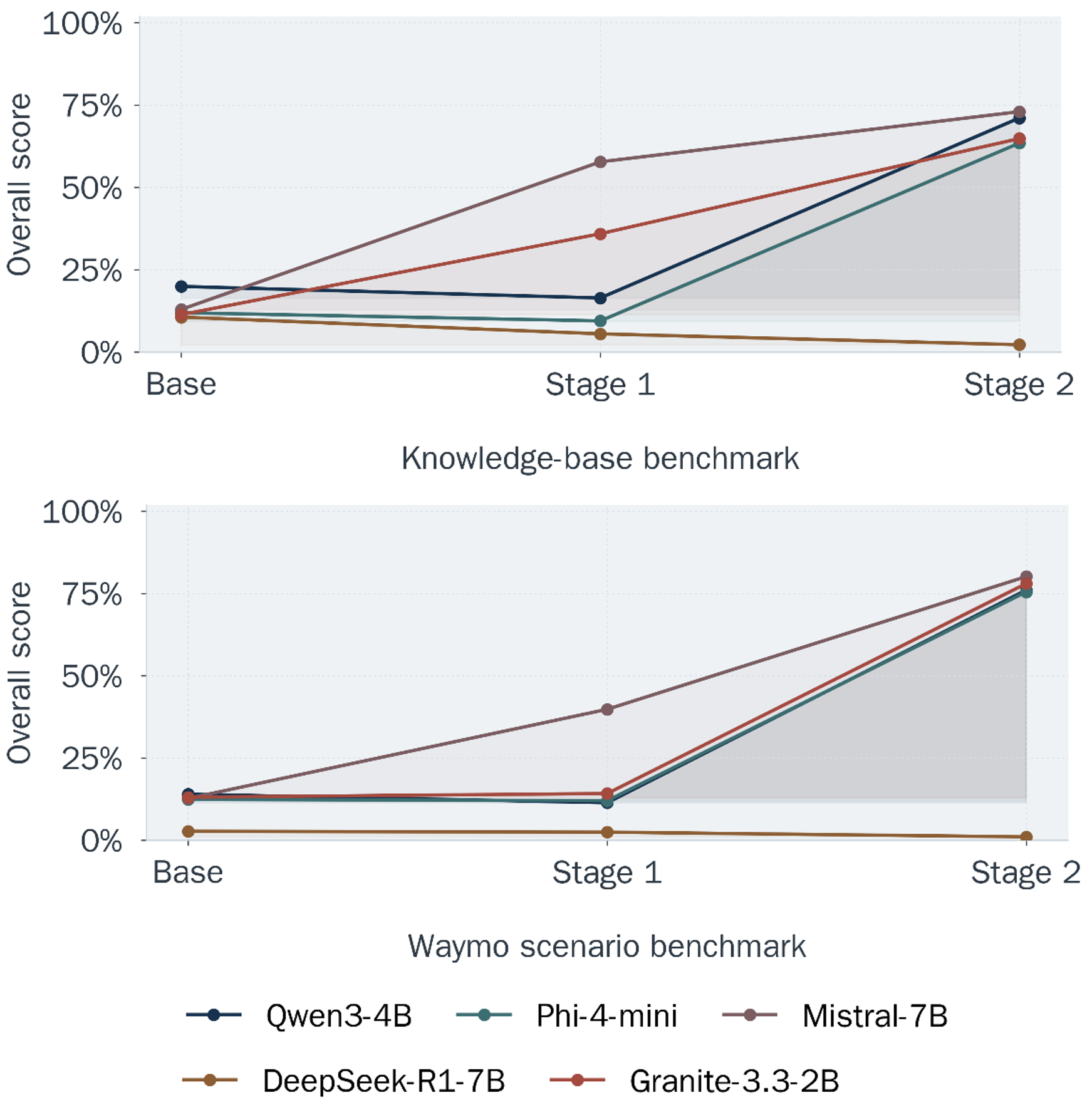}
    \caption{Collision Expert versus base models. The trained Collision Expert achieves more stable structured collision reasoning than generic foundation models, especially in collision-mode recognition, intent inference, and top-k semantic matching.}
    \label{fig_expert_vs_base}
\end{figure}

As shown in Fig.~\ref{fig_expert_vs_base}, the trained Collision Expert provides more reliable structured collision reasoning than generic base models. Although some base models can produce valid structured outputs, they still make substantial errors in semantic fields such as top-1 scenario, top-1 intent, and top-3 hit. In contrast, the Collision Expert is more stable in output validity, collision-mode recognition, and intent inference. This confirms that KG-ASG does not rely on generic foundation-model prompting alone, but requires collision-knowledge alignment and generation-oriented adaptation.

Valid Primary Attack requires that the final ego collision is triggered by the primary adversary, while no support collision, multi-collision, road violation, or signal violation occurs. To jointly evaluate attack success and collision-attribution quality, we define QASR as
\begin{equation}
\begin{aligned}
\mathrm{QASR}
= {} &
\mathrm{ASR}
\cdot
\mathrm{PrimaryOnly}  \\
& \cdot
(1-\mathrm{SupportVio})
\cdot
(1-\mathrm{MultiVio}) .
\end{aligned}
\end{equation}
Here, \(\mathrm{PrimaryOnly}\) indicates whether the final collision is caused by the primary adversary, while \(\mathrm{SupportVio}\) and \(\mathrm{MultiVio}\) denote support-vehicle collision violation and multi-collision violation, respectively.

\begin{table}[t]
\centering
\caption{Expert Selection Accuracy and Constrained Attack Quality}
\label{tab_expert_ablation}
\footnotesize
\renewcommand{\arraystretch}{1.28}
\setlength{\tabcolsep}{3.0pt}
\begin{tabularx}{\columnwidth}{@{}p{0.25\columnwidth}|YYYY@{}}
\toprule
Variant &
\makecell{Raw\\ASR \(\uparrow\)} &
\makecell{Primary\\Match \(\uparrow\)} &
\makecell{Valid\\Primary\\Attack \(\uparrow\)} &
\makecell{QASR\\\(\uparrow\)} \\
\midrule
Expert-Stage1 &
\underline{92.60\%} &
89.40\% &
\underline{84.20\%} &
\underline{77.97\%} \\

Expert-Stage2 &
\textbf{92.80\%} &
\underline{90.60\%} &
68.80\% &
52.89\% \\

KG-ASG Full &
\underline{92.60\%} &
\textbf{98.80\%} &
\textbf{92.20\%} &
\textbf{85.38\%} \\
\bottomrule
\end{tabularx}
\end{table}

Table~\ref{tab_expert_ablation} shows that Raw ASR alone does not fully reflect attack quality. Expert-Stage2 obtains the highest Raw ASR of 92.80\%, but its Valid Primary Attack decreases to 68.80\%, indicating that many successful attacks are accompanied by invalid attribution or multi-collision. In contrast, KG-ASG Full maintains a comparable Raw ASR of 92.60\%, while improving Primary Match to 98.80\%, Valid Primary Attack to 92.20\%, and QASR to 85.38\%. These results show that the full model does not simply increase collision frequency, but converts successful attacks into attributable, interpretable, and single-collider-compliant adversarial scenarios.

\begin{figure}[t]
\centering
\includegraphics[width=\columnwidth]{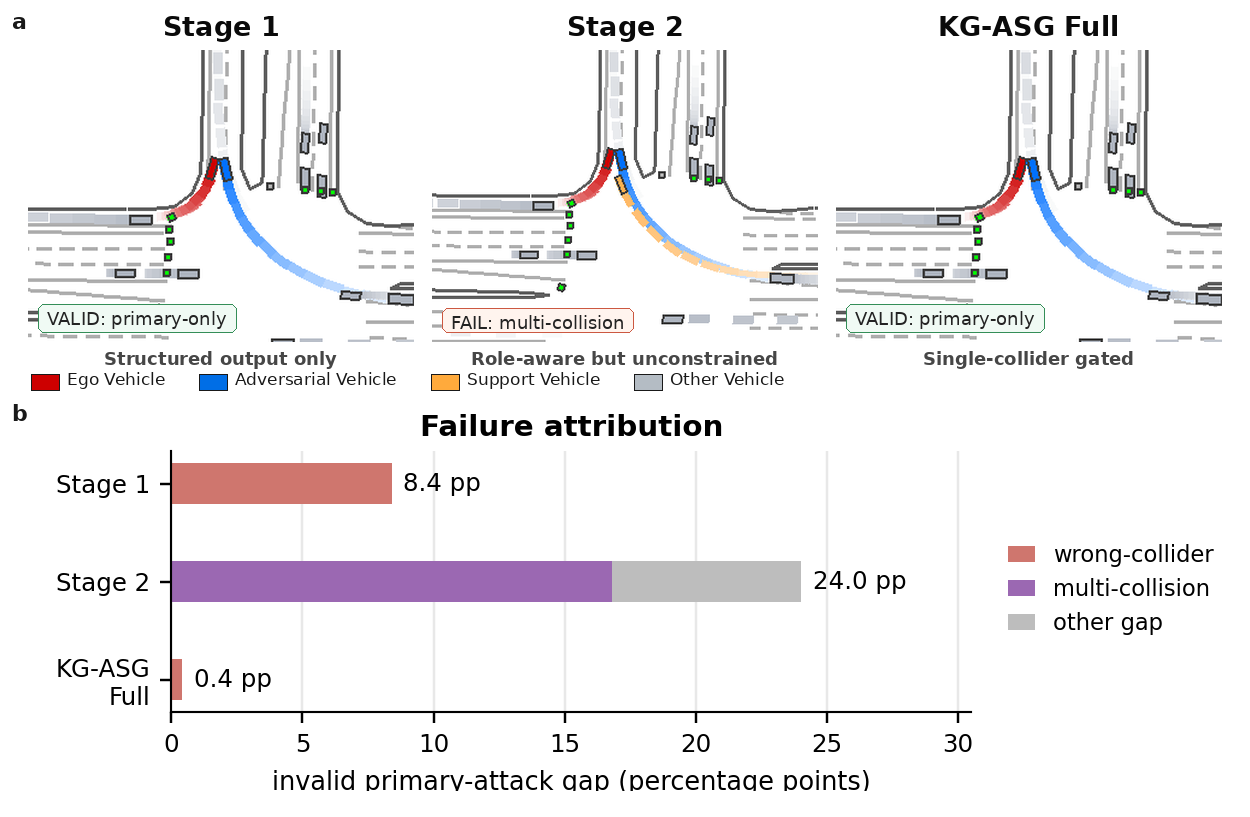}
\caption{Qualitative progression from Stage 1 to KG-ASG Full in failure correction. Stage 1 may produce wrong-collider failures, where the final collision is not caused by the intended primary adversary. Stage 2 strengthens role-aware generation, but may still suffer from multi-collision due to the lack of strict single-collider control. KG-ASG Full resolves these failure modes by enforcing primary-only collision and support no-collision constraints, thereby converting the scenario into a valid primary-only attack. The lower panel further visualizes the invalid primary-attack gap across different stages, showing that KG-ASG Full substantially reduces attribution errors and multi-collision failures.}
\label{fig:expert_failure_progression}
\end{figure}

Fig.~\ref{fig:expert_failure_progression} visualizes how invalid primary attacks are corrected across stages. Stage 1 already produces structured collision semantics, but the final collision can still be caused by a non-primary vehicle. Stage 2 strengthens role-aware generation, but support vehicles may enter the main conflict region and cause multi-collision. KG-ASG Full enforces primary-only collision and support no-collision gates, converting the same interaction into a valid primary-only attack. The lower panel further shows that the invalid primary-attack gap decreases from 8.4 pp in Stage 1 and 24.0 pp in Stage 2 to only 0.4 pp in KG-ASG Full. This qualitative evidence supports Table~\ref{tab_expert_ablation} and shows that KG-ASG improves attribution quality rather than simply increasing Raw ASR.

\begin{table}[t]
\centering
\caption{No-Expert Same-Budget Comparison}
\label{tab_no_expert_same_budget}
\footnotesize
\renewcommand{\arraystretch}{1.20}
\setlength{\tabcolsep}{5pt}
\begin{tabular}{l|cc}
\toprule
Variant & Raw ASR \(\uparrow\) & Multi-collision \(\downarrow\) \\
\midrule
No Expert Same Budget & \textbf{93.00\%} & 43.60\% \\
KG-ASG Full & \underline{92.60\%} & \textbf{0.00\%} \\
\bottomrule
\end{tabular}
\end{table}

To further verify that this improvement does not simply come from allowing more adversarial vehicles, we compare KG-ASG Full with a no-expert variant under the same multi-vehicle budget. As shown in the Table~\ref{tab_no_expert_same_budget}, under the same candidate shortlist and a budget of up to three adversarial vehicles, No Expert Same Budget achieves a slightly higher Raw ASR of 93.00\%, but 43.60\% of its samples contain multi-collision. This leads to unclear collision responsibility and weakens scenario interpretability. By contrast, KG-ASG maintains a comparable Raw ASR while reducing multi-collision to 0.00\%. This demonstrates that the effectiveness of KG-ASG comes from collision-knowledge-guided primary-support selection and single-collider constraints, rather than from unconstrained use of more adversarial vehicles.

\subsection{Closed-Loop Feedback Evolution and Qualitative Analysis}

Open-loop generation evaluates collisions against replayed or predicted ego trajectories, whereas closed-loop evaluation allows the planner-controller to brake, steer, or adjust timing during rollout. Therefore, closed-loop ASR is usually lower than open-loop ASR and should not be directly compared with open-loop attack rates. To evaluate feedback regeneration fairly, each planner-controller is compared with its own closed-loop baseline under the same scene set, controller type, and initial generated scenarios. Recovery Gain is computed using paired scene-level comparison, and 95\% confidence intervals are estimated through bootstrap resampling over scenarios.

\begin{table}[t]
\centering
\caption{Closed-Loop Recovery Under Planner-Controller Feedback}
\label{tab_closed_loop_recovery}
\footnotesize
\renewcommand{\arraystretch}{1.28}
\setlength{\tabcolsep}{2.6pt}
\begin{tabularx}{\columnwidth}{@{}p{0.17\columnwidth}|YYYY@{}}
\toprule
Controller &
\makecell{CL-ASR\\\(N=1\) \(\uparrow\)} &
\makecell{CL-ASR\\\(N=5\) \(\uparrow\)} &
\makecell{Recovery\\Gain \(\uparrow\)} &
\makecell{Failed-set\\Recovery \(\uparrow\)} \\
\midrule
IDM &
\textbf{47.40\(\pm\)4.50} &
\textbf{55.40\(\pm\)4.40} &
\underline{+8.00\(\pm\)2.50} &
\underline{34.91\(\pm\)8.99} \\

Cruise &
\underline{44.60\(\pm\)4.40} &
51.80\(\pm\)4.40 &
+7.20\(\pm\)2.31 &
33.88\(\pm\)8.43 \\

Expert &
39.00\(\pm\)4.30 &
\underline{53.00\(\pm\)4.40} &
\textbf{+14.00\(\pm\)3.20} &
\textbf{52.55\(\pm\)8.39} \\
\bottomrule
\end{tabularx}

\vspace{1.0em}
\begin{minipage}{0.98\columnwidth}
\footnotesize
\emph{Note:} The values after \(\pm\) denote half-widths of 95\% scene-level paired bootstrap confidence intervals estimated over 500 WOMD/MetaDrive scenarios with 10,000 resamples. \(N=1\) denotes closed-loop evaluation without feedback retry, while \(N=5\) enables up to five controller-conditioned retry rounds.
\end{minipage}
\end{table}

As shown in Table~\ref{tab_closed_loop_recovery}, planner-controller feedback consistently improves closed-loop attack recovery across all tested controllers. Compared with \(N=1\), enabling up to five feedback-conditioned retry rounds increases CL-ASR from 47.40\% to 55.40\% for IDM, from 44.60\% to 51.80\% for Cruise, and from 39.00\% to 53.00\% for the Expert controller. The corresponding Recovery Gains are \(+8.00\pm2.50\), \(+7.20\pm2.31\), and \(+14.00\pm3.20\) percentage points, respectively. Since the lower bounds of the 95\% paired bootstrap confidence intervals are all above zero, the improvement is stable under scene-level resampling rather than being caused by incidental fluctuations. All results are reproduced under the CAT code framework. During closed-loop retry, the planner-controller remains fixed and is not trained, retry parameters are taken from the validation configuration, and support no-collision and primary-only constraints remain active throughout the process. Therefore, the recovery gain mainly comes from feedback-driven correction of conflict timing, terminal position, and candidate ranking, rather than from relaxing attribution constraints or modifying the tested controller.

\begin{figure*}[t]
    \centering
    \includegraphics[width=\textwidth]{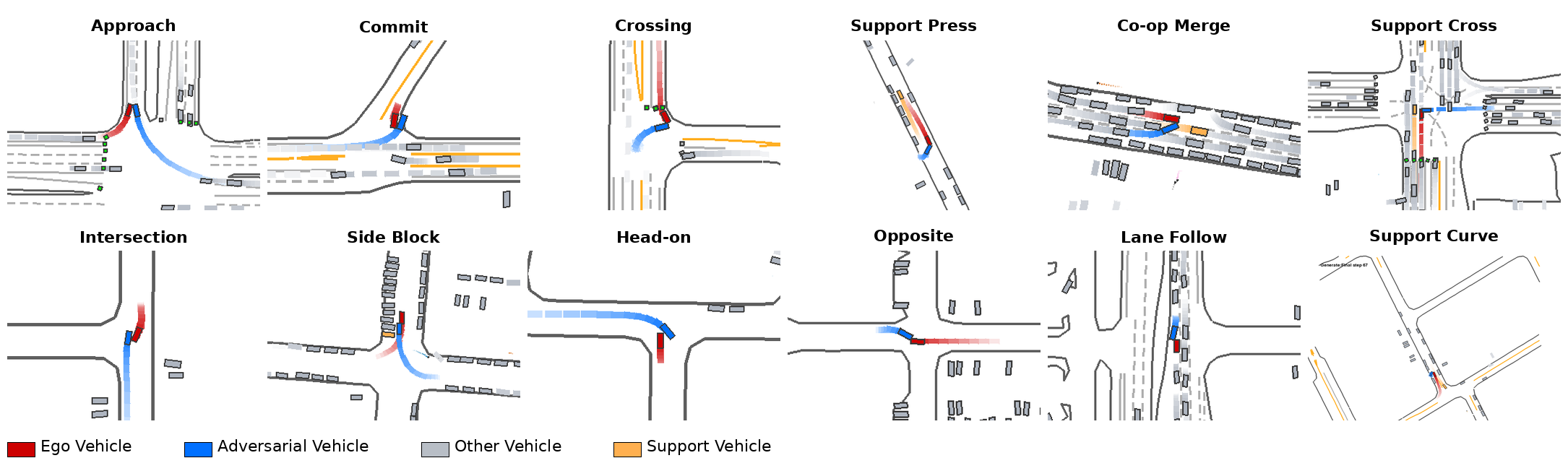}
    \caption{Multi-modal KG-ASG cases with primary-support roles. KG-ASG generates diverse high-risk interaction structures, where the primary adversary is responsible for the final conflict and support or pressure vehicles shape the risk structure without becoming final colliders.}
    \label{fig_multimodal_cases}
\end{figure*}

\begin{figure*}[t]
    \centering
    \includegraphics[width=\textwidth]{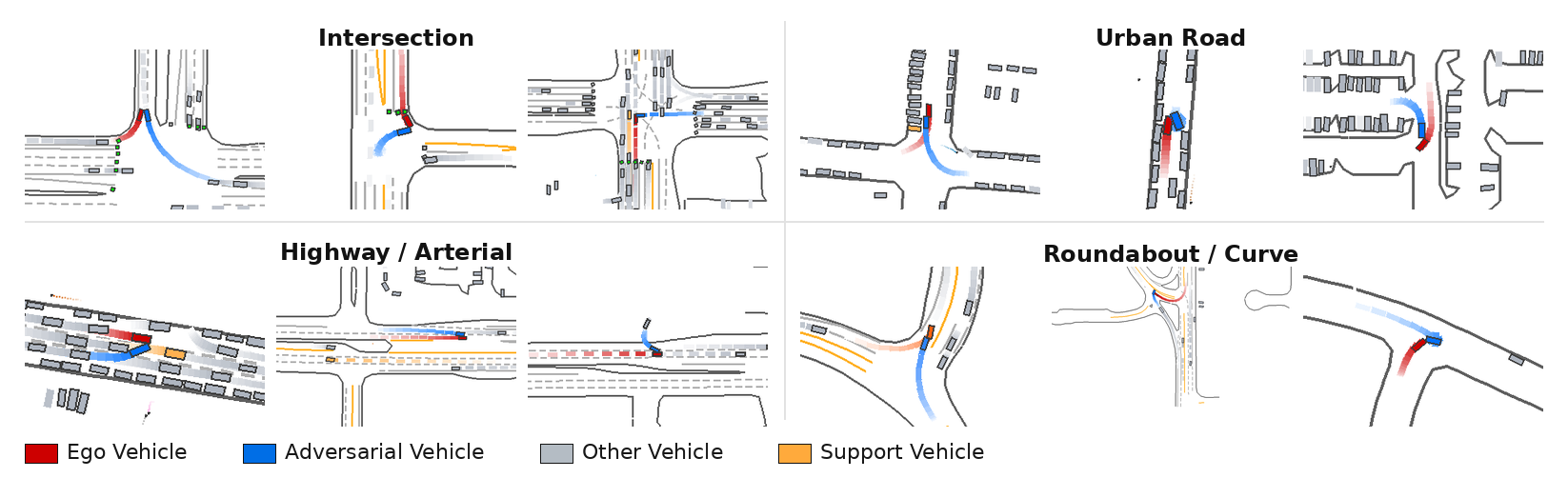}
    \caption{Scene-type stratification of KG-ASG generated scenarios. The generated scenarios are grouped according to traffic structure, lane relation, collision intent, and target conflict zone. The distribution shows that KG-ASG covers heterogeneous interaction patterns rather than relying on a single rear-end, cut-in, or intersection-conflict template.}
    \label{fig_scene_type_stratification}
\end{figure*}

Figs.~\ref{fig_multimodal_cases} and~\ref{fig_scene_type_stratification} jointly illustrate the risk diversity of KG-ASG. The generated scenarios cover multiple road structures and interaction types, including intersections, urban roads, highway or arterial segments, and roundabout or curved-road cases. This indicates that KG-ASG is not tailored to a specific road type or collision pattern, but can generate diverse, attributable, and primary-support-structured risk scenarios across heterogeneous traffic contexts.

Overall, the experiments show that KG-ASG improves safety-testing value in three aspects rather than merely increasing ASR. First, it maintains strong attack effectiveness while reducing irrelevant-object collisions and line-crossing violations. Second, it converts raw collision success into valid primary-only attacks with explicit collision attribution. Third, it recovers attacks avoided by reactive planner-controllers without relaxing rule, physical, or support no-collision constraints. These results indicate that KG-ASG generates scenarios that are not only adversarial, but also executable, attributable, and suitable for reproducible autonomous-driving safety validation.

\section{Conclusion}

This paper proposed KG-ASG, a knowledge-guided closed-loop adversarial scenario generation framework for safety testing of autonomous-driving planner-controller systems. Unlike generation methods that rely mainly on low-level collision proxies or unconstrained multi-vehicle perturbations, KG-ASG integrates collision-knowledge reasoning, primary-support actor selection, rule-constrained trajectory generation, and planner-controller feedback regeneration into a unified framework. Specifically, KG-ASG first constructs a structured collision knowledge base and trains a lightweight Collision Expert to infer the target collision mode, the unique primary adversary, support-vehicle roles, and structured generation guidance. It then formulates multi-vehicle risk generation as a primary-support process, where the primary adversary induces the main conflict and support vehicles shape the feasible avoidance pressure. Road, signal, dynamic, interaction-safety, and single-collider constraints are further embedded as hard gates to prevent adversarial samples from degenerating into unrealistic or non-executable collisions. Finally, planner-controller feedback is used to retry failed samples, re-rank candidates, and refine terminal trajectories, thereby improving attack recovery against reactive controllers. \par

Experimental results show that KG-ASG preserves strong open-loop adversarial capability while substantially improving interpretability and executability of generated scenarios. Under the CAT-compatible protocol, KG-ASG achieves nearly 93\% attack success rate under the Replay policy. Factorized ablation results show that rule constraints effectively suppress irrelevant-object collisions, road departures, and signal violations, while physical constraints significantly reduce acceleration violations. Their combination maintains a high attack success rate while improving scenario validity. Further primary-support ablations demonstrate that simply increasing the number of adversarial vehicles can increase Raw ASR but causes severe multi-collision artifacts. In contrast, KG-ASG maintains comparable Raw ASR, reduces multi-collision to 0.00\%, and substantially improves Valid Primary Attack and QASR. Closed-loop experiments further validate the effectiveness of planner-controller feedback regeneration, where KG-ASG obtains consistent recovery gains under IDM, Cruise, and Expert controllers. \par

Future work will proceed in three directions. First, the collision knowledge base will be expanded with more real-world pre-crash states, disengagement events, and multi-agent interaction patterns to improve generalization to complex long-tail conflicts. Second, support-vehicle role modeling will be refined to better distinguish fine-grained risk-shaping behaviors, including blocking, rear pressure, merge interference, and crossing-triggered conflicts. Third, we will extend KG-ASG to additional ODDs, larger-scale mixed-traffic scenarios, and sensor-level closed-loop testing to further evaluate cross-platform generalization. \par


\ifCLASSOPTIONcaptionsoff
  \newpage
\fi

\bibliographystyle{IEEEtran}
\bibliography{re}

\vspace{11pt}

\end{document}